# Product Aesthetic Design:
# A Machine Learning Augmentation

by

Alex Burnap

John R. Hauser

and

Artem Timoshenko

November 2022

Alex Burnap is an Assistant Professor of Marketing at Yale School of Management, Yale University, 165 Whitney Avenue, Evans Hall 5467, New Haven, CT 06511, (405) 880-3660, alex.burnap@yale.edu.

John R. Hauser is the Kirin Professor of Marketing, MIT Management School, Massachusetts Institute of Technology, E62-538, 77 Massachusetts Avenue, Cambridge, MA 02139, (617) 253-2929, hauser@mit.edu.

Artem Timoshenko is an Assistant Professor of Marketing at Kellogg School of Management, Northwestern University, 2211 Campus Drive, Suite 5391, Evanston, IL 60208, (617) 803-5630, artem.timoshenko@kellogg.northwestern.edu.

We thank Jeff Hartley, John Manoogian II, Andrew Norton, Joyce Salisbury, Zheng Shen, and Sharon Sheremet for valuable insights into how product aesthetics are designed and evaluated; Mark Beltramo, Fred Feinberg, Ari Helljaka, Honglak Lee, Ye Liu, and Yanxin Pan for mathematical modeling discussion; and Emrah Bayrak, Songting Dong, Dean Eckles, Nasreddine El Dehaibi, Siham El Kihal, Gui Liberali, Erin MacDonald, Ye Liu, Max Yi Ren, and Glen Urban for helpful comments and suggestions.

# Product Aesthetic Design:
# A Machine Learning Augmentation

## Abstract

Aesthetics are critically important to market acceptance. In the automotive industry, an improved aesthetic design can boost sales by 30% or more. Firms invest heavily in designing and testing aesthetics. A single automotive "theme clinic" can cost over $100,000, and hundreds are conducted annually. We propose a model to augment the commonly-used aesthetic design process by predicting aesthetic scores and automatically generating innovative and appealing product designs. The model combines a probabilistic variational autoencoder (VAE) with adversarial components from generative adversarial networks (GAN) and a supervised learning component. We train and evaluate the model with data from an automotive partner—images of 203 SUVs evaluated by targeted consumers and 180,000 high-quality unrated images. Our model predicts well the appeal of new aesthetic designs—43.5% improvement relative to a uniform baseline and substantial improvement over conventional machine learning models and pretrained deep neural networks. New automotive designs are generated in a controllable manner for use by design teams. We empirically verify that automatically generated designs are (1) appealing to consumers and (2) resemble designs which were introduced to the market five years after our data were collected. We provide an additional proof-of-concept application using opensource images of dining room chairs.

## 1. Introduction

Consumers consistently rank aesthetics among the three most important factors in product choice (Bloch 1995; Creusen and Schoormans 2005). For example, the visual design of the original iPod was judged to be a critical factor in its market acceptance (Reppel, Szmigin, and Gruber 2006). In categories such as home appliances, aesthetics help firms establish product differentiation beyond functional characteristics (Bloch 1995; Crilly, Moultrie, and Clarkson 2004; Person et al. 2007); for instance, the Dyson DC01 used transparent design to communicate its complexity to consumers, helping it become the best-selling vacuum in the U.K. (Noble and Kumar 2010). Firms use aesthetics to strategically position and enhance brand recognition (Aaker and Keller 1990; Karjalainen and Snelders 2010; Keller 2003). Trade dress violations (non-functional attributes that signal brand identity) are hotly contested in Lanham Act (§43A) litigation. Aesthetics pervade marketing—visually-appealing products and packaging drive consumers to choose one product over another, especially at the point of purchase in crowded brick-and-mortar stores, supermarkets, and online retailers (Clement 2007; Orth and Malkewitz 2008).

Developing product aesthetics can require substantial investment, yet returns on investment are found across markets—a study of 93 firms across nine product categories found that firms that heavily invested in aesthetic design had 32% higher earnings than industry averages (Hertenstein, Platt, and Veryzer 2005). Marketing and product managers routinely manage the aesthetic design of products, services, and digital marketplaces. In this paper, we propose a methodology to improve the process of aesthetic product design and testing. The basic concepts of the proposed methodology are applicable across product categories. Our research focuses on the automotive industry where we have the most experience and an industry partner; we provide an additional proof-of-concept application using publicly available data on furniture.

In the automotive industry, product aesthetics explain up to 60% of consumers' purchase decisions (Kreuzbauer and Malter 2005). Automotive design significantly affects market performance (Cho, Hasija, and Sosa 2015; Jindal et al. 2016; Rubera 2015), in large part by influencing consumer consideration (Liu et al. 2017; Palazzolo and Feinberg 2015). For example, the redesign of the 2008 Buick Enclave commanded a 30% increase in MSRP over the Buick Rendezvous it replaced (using the same engine; Figure 1); the redesign of the 2005 Volkswagen Beetle resulted in a 54% market share gain in a single year (Kreuzbauer and Malter 2005; Blonigen, Knittel, and Soderbery 2013). On the other hand, the aesthetics of the 2001 Pontiac Aztec was cited as a primary reason for its market failure (Vlasic 2011). Not surprisingly, automotive firms invest heavily in design—$1.25 billion on average per model, and up to $3 billion for major redesigns involving both styling and platform (Blonigen, Knittel, and Soderbery 2013; Pauwels et al. 2004; Rubera 2015).

**Figure 1.** Illustrative Example of Three Otherwise Similar Automobiles with Different Aesthetic Design.

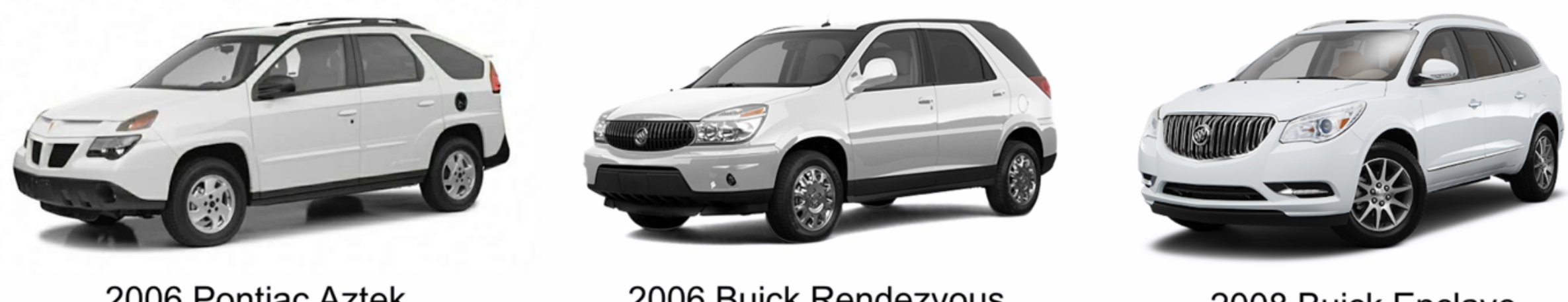

Traditionally, human judgment drives the aesthetic design in at least two ways. First, while there are established aesthetic heuristics and cognitive design principles (Coates 2003; Crilly, Moultrie, and Clarkson 2004; Norman 2004), aesthetic design is often generated and screened by design teams who have an "eye" for visual design. Design teams are powerful within organizations; their aesthetic judgments are hard to overrule (Vlasic 2011).

Human judgment also affects aesthetic design through consumer evaluations. Firms often ask consumers to evaluate alternative designs in laboratory test markets, A/B tests, or "theme clinics." In a typical automotive theme clinic, a few hundred targeted consumers are recruited and brought to a central location to evaluate aesthetic designs. Consumers view the aesthetic designs and rate them on established benchmarks such as semantic scales for sporty, appealing, innovative, and luxurious (Coates 2003; Manoogian II 2013). Theme clinics are costly. Automotive firms typically invest over $100,000 per theme clinic for a single new vehicle design. With multiple aesthetic designs per vehicle and over a hundred vehicles in its worldwide product line, General Motors alone spends tens of millions of dollars on theme clinics. With additional costs incurred when designers screen hundreds of aesthetic designs down to those destined for theme clinics, annual costs can exceed $100 million for a single manufacturer.

We propose methods to augment the traditional product development process with machine learning tools to address both aspects of aesthetic design: (1) the generation and (2) the testing of new aesthetic designs. For testing, the model predicts how consumers would rate aesthetic designs directly from visual images. We demonstrate that the model can predict different semantic scales, such as aesthetic appeal, innovativeness, or traditional vs. modern. Specifically, we use an *encoding model* to represent visual designs (images) using 512-dimensional embeddings, and train a *predictive model* that predicts aesthetic ratings based on those embeddings. The predictive model is designed to screen newly proposed aesthetic designs so that only the highest-potential designs need to be tested in theme clinics (or their equivalent for non-automotive applications).

For generation, the *generative model* creates new product designs with attributes defined by the design team (e.g., "Cadillac-like"). This gives the designers a tool to morph through the design space and explore visual dimensions of consumer aesthetic perceptions. These generated designs can be rated by the *predictive model* to identify those with high aesthetic scores. In the automotive proof-of-concept, we demonstrate that generated images are controllable, realistic, and perceived by consumers as aesthetically appealing. Moreover, the model can be "creative": When trained using model year 2010-2014 data, the model can generate automotive designs similar to those introduced in model year 2020.

Our research was influenced by senior design and marketing managers in the automotive industry interested in using machine learning methods to improve aesthetic product design. These managers suggested that we focus on *augmentation* of human expertise and creativity in the traditional aesthetic design workflow, rather than *automation*. Design teams welcome augmentation, but organizational structure, history, and designers' beliefs and training resist full automation (Coates 2003). Our experiences with real design teams guide the modeling decisions in the proposed approach.

## 2. Conceptual Model of Product Aesthetic Design

### 2.1. Augmenting the Design Process

Figure 2 summarizes the current widely-used design process and the proposed machine learning augmentation. Consider first the current process as shown in the first two rows (coded as black). The process begins with a market definition that is external to the aesthetic design efforts. For example, Apple targeted smartphones with a touchscreen; Zenni Optical aimed to develop prescription sports glasses; IKEA targeted affordable yet aesthetically pleasing furniture. In automotive markets, firms target particular segments such as luxury compact utility vehicles (currently the Cadillac XT5, Buick Envision, Volvo XC60, BMW X3, and others). Market definition provides soft constraints on aesthetics based on the targeted consumers and the firm's capabilities (Box 1).

Aesthetic designers create hundreds to thousands of freehand sketches that are converted to 2D images (Box 2; Coates 2003). For example, Dyson and General Motors generate several hundred sketches per new device or vehicle, while IKEA can generate fewer sketches given its product line variety and turnover (Bouchard, Aoussat, and Duchamp 2006; Toffoletto 2013). The human design team next screens potential designs to a smaller set of testable design concepts in a process known in design as "down-selection" (Box 3). Consumers evaluate the testable designs in theme clinics resulting in more screening. Successful designs are advanced downstream for further development, including engineering, manufacturing, and marketing communications (advertising, social media, websites). The process is highly iterative and asynchronous across both design concept generation and testing.

**Figure 2.** Augmenting Aesthetic Design with Machine Learning

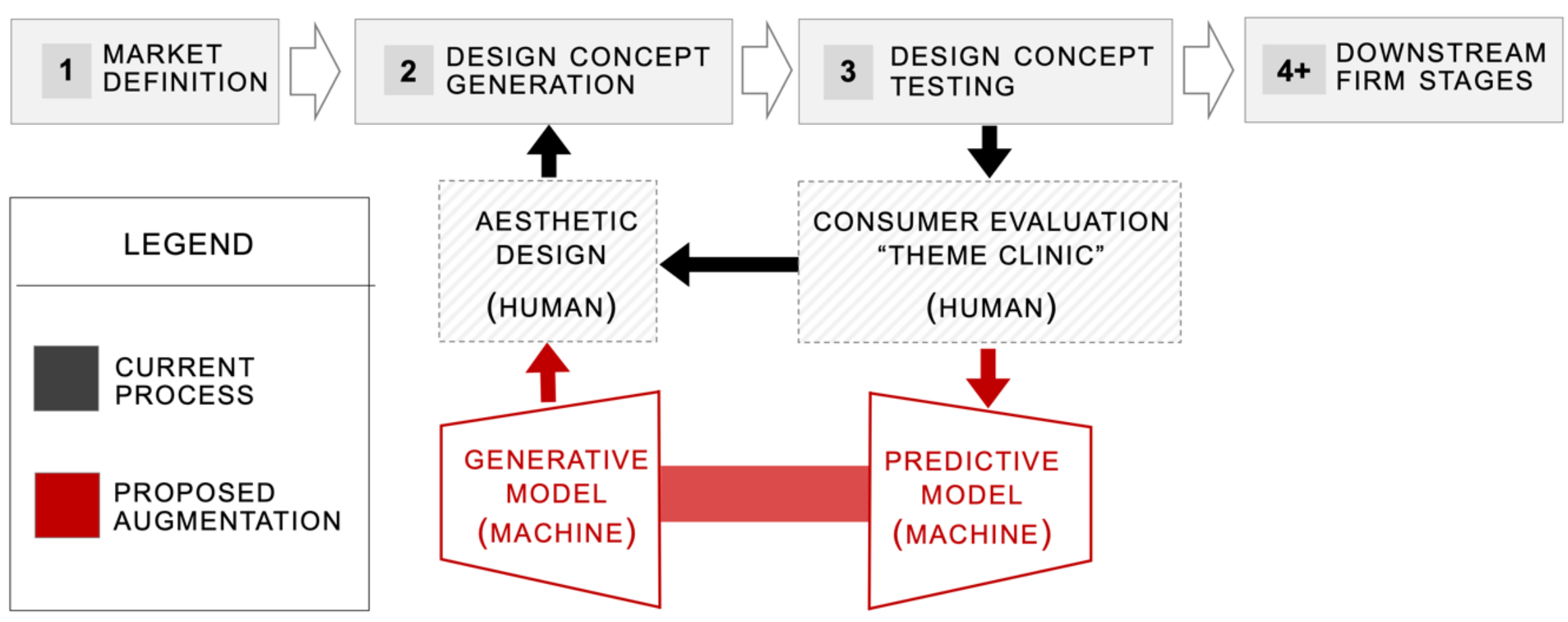

The red trapezoids and arrows highlight the proposed machine learning augmentation. The machine learning models augment the traditional design process and apply to all iterations in concept generation and testing. In testing, the *predictive model* helps eliminate designs likely to score low in theme clinics. Focusing on high-potential product images improves the traditional design workflow in several ways. Accurate prediction provides quick feedback and enables faster iterations by the design team. Theme clinics also become more efficient and effective because less respondent time is allocated to images with low predicted scores (Gross 1972). As a result, firms benefit from shorter product development times and cost reductions due to a lower product "drop rate," i.e., the share of design concepts later terminated in downstream stages (Cooper 1990; Danneels and Kleinschmidtb 2001). Finally, rigorous quantitative evaluation helps "shield" aesthetic designs from downstream changes driven by engineering, manufacturing, or accounting (Hartley 1996; Manoogian II 2013; Vlasic 2011).

The *generative model* creates designs that are realistic and screened by the *predictive model* to be aesthetically pleasing. Generated designs are intended to spark creativity among human designers, who can use the model as a tool to explore the space of possible aesthetic designs (Martindale 1990). The designers "control" the generative model through specifying attributes. Example attributes are 'red,' 'Cadillac-like,' 'Sport Utility Vehicle (SUV),' '2015 vintage,' or 'viewed from the side.' The vintage variable is important when training the model in our empirical application. It captures the evolving aesthetics in the automotive industry (Hekkert, Snelders, and Wieringen 2003; Martindale 1990).

## 2.2. Technical and Managerial Challenges in Augmenting Aesthetic Design

The efficient augmentation of the traditional aesthetic design pipeline with machine learning tools requires that we address technical and managerial challenges. First, images pose a technical challenge as they are inherently high dimensional. Even modest quality images are 100 x 100 pixels for each of red, green, and blue colors together comprising 30,000 variables—far too many to be input to conventional choice models. Previous work has represented aesthetics in choice models using hand-engineered features such as characteristic lines (Chan, Mihm, and Sosa 2018; Ranscombe et al. 2012), landmark points (Landwehr, Labroo, and Herrmann 2011), silhouettes (Orsborn, Cagan, and Boatwright 2009; Reid, Gonzalez, and Papalambros 2010), and Bezier curves (Kang et al. 2016); or aggregated numbers such as J.D. Power APEAL or online reviews (Cho, Hasija, and Sosa 2015; Homburg, Schwemmle, and Kuehnl 2015; Jindal et al. 2016; Pauwels et al. 2004). Despite the challenges of working with images, we follow the industry standard providing realism to designers who think naturally in terms of holistic images. Images are, to designers and consumers, realistic representations of new product aesthetics.[1]

Second, gathering consumer evaluations is costly and results in limited training data. In our automotive application, we are fortunate to have 7,308 aesthetic ratings by consumers for 203 vehicles, but those ratings alone would be insufficient to estimate a predictive and/or generative model with high-dimensional image data.

We address these practical challenges by training an *encoding model* to embed images in a lower-dimensional vector space. Embeddings reduce the dimensionality of the images for the *predictive* and *generative models* by leveraging both the relatively thin and expensive labeled training data (images with aesthetic ratings) with a much larger sample of unlabeled training data (180,000 images without consumer evaluations).[2] Success depends upon whether the embeddings compress the important information from the full images while allowing us to predict human aesthetic judgments and generate perceptually realistic new designs.

Embeddings using a neural network have seen recent adoption in marketing science. For example, Timoshenko and Hauser (2019) embed textual data to identify consumer needs; Liu, Lee, and Srinivasan (2019) embed product reviews to predict sales conversion; Liu, Dzyabura, and Mizik (2020) embed social media images to predict identity; Dew, Ansari and Toubia (2022) embed firms' logos to describe brand personality and similarity; Gabel and Timoshenko (2022) embed purchase histories to predict product choice in retail; and Chakaborthy, Kim and Sudhir (2019) embed reviews to identify sentiment and missing evaluations. In this paper, we embed product images to predict aesthetic ratings and generate new aesthetic designs.

[1] Even without working at the pixel level, aesthetic design is high dimensional. Pfitzer and Rudolph (2007) describe 17 distinct elements (e.g., doors, roof, pillars, headlights lights) and 10 different design lines (e.g., roof line, cowl line, rail line). Each of these elements has multiple levels. All elements interact for visual and emotional appeal.

[2] In our automotive application, the images contain 786,432-dimensions based on 512 x 512 x 3 (height x width x color) pixels. The dimensionality of the embedding space (512-dimensional) is a fine-tuning decision (§5.2).

Third, aesthetic evaluations are holistic (Berlyne 1971; Bloch 1995; Crilly, Moultrie, and Clarkson 2004; Martindale 1990). Design aspects are interdependent; we cannot expect consumers to evaluate the design aspects separately as would be done in conjoint analysis (Orme and Chrzan 2017). For example, a consumer cannot evaluate the aesthetics of a new BMW X3 design as an additive sum of the shape and position of headlights, the slope of the hood, and the height of the beltline. Rather, the Gestalt interplay of all design elements, including subtle elements such as the "Hoffmeister kink," drive consumer evaluations of qualitative attributes such as appealing, sporty, aggressive, luxurious, or modern (Coates 2003). By using deep neural networks for the encoding, predictive, and generative models, we automatically and holistically capture the interplay of aesthetic elements.

Fourth, the aesthetic design process is highly iterative, asynchronous, and distributed. This poses a significant managerial challenge—multiple design teams (and sub-teams) must be able to use the machine learning augmentation for their corresponding roles in design concept testing and generation. Integrating machine learning into the existing aesthetic design process must delicately balance its interplay with the established workflows. For example, the design team may split into sub-teams to work on several promising design concepts in parallel, while concurrent theme clinics may be testing entirely different design concepts. To enhance parallel development, once trained, our proposed predictive and generative models can be used independently or together as needed.

## 3. Overview of a Machine Learning Approach to Augment Aesthetic Product Design

Let $X_i$ be the (height x width x color) 3D tensor of pixels of image $i$. Our model requires two inputs: product images labeled with aesthetic ratings evaluated by consumers, $\{(X_i, y_i)|\ i = 1..N_L\}$, and unlabeled product images without ratings but with product attributes, $\{(X_i, \vec{a}_i)|\ i = 1..N_U\}$. For example, $y_i$ might be the average consumer rating of the aesthetic appeal of product design $i$, and the attributes $\vec{a}_i$ may describe primary exterior color, body type, model year, or brand. Firms typically obtain consumer evaluations for a small fraction of images, $N_L \ll N_U$. Attributes are important for managerial acceptance because they enable the design team to control the design. However, the proposed model does not require that all attributes are available for every image. The model imputes any missing attributes during inference (§4).

Our two high-level goals are (1) to test new product aesthetics by predicting consumer ratings,

$\hat{y}_{new}$, for new product images created by the design teams, $X_{new}$, and (2) generate new product designs, $\hat{X}_{gen}$, according to attributes desired by the design team, $\vec{a}_{gen}$, such that images score well on ratings, $\hat{y}_{gen}$. We summarize notation in Appendix A.1.

**Figure 3.** Proposed Machine Learning Augmentation Model

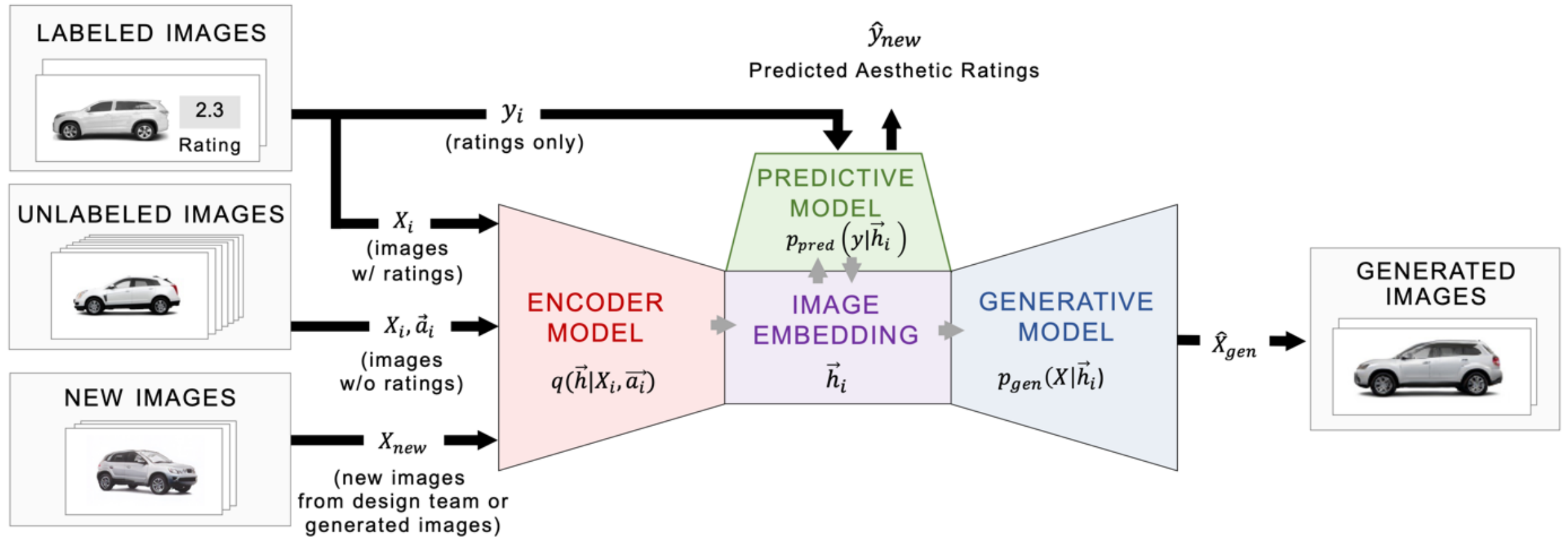

Figure 3 illustrates the general input and output flow of the proposed machine learning augmentation for aesthetic design. For every aesthetic design $i$, the encoder model inputs the image, $X_i$, and/or attributes, $\vec{a}_i$, and outputs a 512-dimensional embedding distribution $q(\vec{h}|X_i, \vec{a}_i)$. We sample the embedding vector $\vec{h}_i$ from the distribution $q(\vec{h}|X_i, \vec{a}_i)$. The predictive model uses $\vec{h}_i$ to predict aesthetic scores, $\hat{y}_i = p_{pred}(y|\vec{h}_i)$. When predicting the aesthetic rating for a new image $X_{new}$, we use the encoder to obtain an embedding distribution $q_{new}(\vec{h}) = q(\vec{h}|X_{new}, -)$, and we average predictions over multiple draws of the embedding vectors, $\hat{y}_{new} = \mathbb{E}_{\vec{h}_i \sim q_{new}(\vec{h})}[p_{pred}(y|\vec{h}_i)]$. The generative model creates images conditional on the embedding $\vec{h}_i$, meaning $\hat{X}_i = p_{gen}(X|\vec{h}_i)$. When generating new aesthetic designs, we input a desired attribute vector $\vec{a}_{gen}$ into the encoder to obtain the distribution $q_{gen}(\vec{h}) = q(\vec{h}|-, \vec{a}_{gen})$, sample embeddings $\vec{h}_r \sim q_{gen}(\vec{h})$, and generate images $\hat{X}_{gen} = p_{gen}(X|\vec{h}_r)$ for embeddings with high predicted appeal $p_{pred}(y|\vec{h}_r)$.

The three models in the proposed approach—the generative model, encoding model, and predictive model—are connected by the probabilistic embedding. We learn an embedding *distribution* for each product design rather than a point estimate. Each aesthetic design $i$ (from designers or automatically generated) has its own embedding distribution $\vec{h}_i \sim q(\vec{h}|X_i, \vec{a}_i)$. Learning the parameters of the distribution leverages the variational inference literature (Blei, Kucukelbir, and McAuliffe 2017; Jordan et al. 1999), enabling Bayesian parameter estimation at data sizes otherwise intractable for MCMC sampling.

The shared probabilistic embedding helps to effectively leverage big unlabeled data in training the predictive model. Intuitively, the labeled data from theme clinics alone are too thin to learn a mapping from high-dimensional product images to the aesthetic ratings. The unlabeled data contain information about the product itself (e.g., automobile images have four wheels). Our model relies on this information to train the probabilistic encoder, which then makes the prediction problem feasible with thin data.

We use deep neural networks to parametrize all three models. We combine and jointly minimize loss functions for the three deep neural networks using both labeled and unlabeled images. To be used by real design teams, the generative model produces images perceived as realistic. We gain realism with adversarial training. Adversarial training requires the generator create images the encoder perceives as real, while the encoder seeks to distinguish real from generated images. The end result of the minmax equilibrium leads to images that are realistic. For the remainder of paper, we refer to the predictive model, generative model, and encoding model as the predictor, generator, and encoder, respectively.

## 4. Proposed Approach: Semi-Supervised Variational Autoencoder with Adversarial Terms

We denote the parameters of the predictive model by $\vec{\beta}_P$, the generative model by $\vec{\beta}_G$, the encoding model by $\vec{\beta}_E$, and the combined vector of parameters by $\vec{\beta} = (\vec{\beta}_P, \vec{\beta}_G, \vec{\beta}_E)$. To train the model, we minimize the combined loss function:

(1) $$\mathcal{L}(\vec{\beta}) = \mathcal{L}_{pred}(\vec{\beta}_P) + \mathcal{L}_{gen}(\vec{\beta}_G) + \mathcal{L}_{enc}(\vec{\beta}_E)$$

where $\mathcal{L}_{pred}(\vec{\beta}_P)$, $\mathcal{L}_{gen}(\vec{\beta}_G)$ and $\mathcal{L}_{enc}(\vec{\beta}_E)$ indicate the predictive, generative, and encoding loss functions, respectively. The summation of the loss terms is theoretically justified by the law of conditional probability and the principles of approximate marginalization of the likelihood formulation of the loss functions. However, we weight the various loss functions when training the model. We provide details for the probabilistic formulation of the VAE and for separability of the loss function in Appendix A.2.

### 4.1. Loss Terms for the Predictive, Generative, and Encoding Models

**Predictive Model.** We use a deep neural network, $f_P(\vec{h}_i, \vec{\beta}_P)$, to map embeddings to the rating of interest, say a 1-to-5 rating on aesthetic "appeal." Information about the full images and attributes is summarized in the embeddings. For the predictive model, we define $\hat{y}_i = f_P(\vec{h}_i, \vec{\beta}_P)$ as the predicted rating from the neural network; the loss term minimizes the mean absolute error of predicted versus true ratings. The mean absolute error definition was motivated by our industry application. Our automotive partner traditionally considers the mean absolute error in their analysis of aesthetic ratings. This loss function is consistent with an assumption that the observed ratings are drawn from a Laplace distribution with mean $f_P(\vec{h}_i, \vec{\beta}_P)$ and unit diversity in the probabilistic formulation (see Appendix A.2).

$$\mathcal{L}_{pred}(\vec{\beta}_P) = -\sum_{i \in rated} |y_i - \hat{y}_i| \tag{2}$$

**Generative Model.** We use a deep neural network, $f_G(\vec{h}_i, \vec{\beta}_G)$, to generate an image $\hat{X}_i$ from an embedding $\vec{h}_i$. The loss function for the generative model combines two terms. The first term rewards the quality of image reconstruction. Given an embedding $\vec{h}_i$ for an image $X_i$ (labeled or unlabeled), we want the generative model to produce an image $\hat{X}_i$ that is similar to the original image $X_i$. This assures that the generated images are "vehicle-like."

We use a second term to enhance the generative model using "masks." A mask defines the general shape of a product, say "SUV-like." Masks are matrices with binary values of the same height and width as the product images. The mask's $D$ pixels, $M_i$, indicate the presence of the product in the image. We use standard computer vision tools to create masks for all labeled and unlabeled images, and show an example mask in Appendix A.4. Masks focus the generative model on product designs rather than unrelated information in product images. In the generator, masks are analogous to a $4^{th}$ color (red, green, blue, mask) and predicted by the same deep neural network using the (now-augmented) parameters, $\vec{\beta}_G$.

Although the generative model is used to generate new designs, it is trained on existing images. As detailed in Appendix A.2, the mean absolute error loss function is consistent with the images (and masks) being drawn from a high-dimensional Laplace distribution with mean $f_G(\vec{h}_r, \vec{\beta}_G)$ and unit diversity. If $d$ indexes the now-4D pixels, then the loss function for the generative model becomes:

$$\mathcal{L}_{gen}(\vec{\beta}_G) = -\sum_{i \in rated, unrated} \left\{ \frac{1}{3D} \sum_d |x_{id} - \hat{x}_{id}| + \frac{1}{D} \sum_d |m_{id} - \hat{m}_{id}| \right\} \tag{3}$$

**Encoding Model.** We use a deep neural network, $f_E(X_i, \vec{a}_i, \vec{\beta}_E)$, to map images, $X_i$, and product attributes, $\vec{a}_i$, to an embedding distribution, $\hat{q}_{enc}(\vec{h} | X_i, \vec{a}_i)$. We assume a $K$-dimensional Gaussian distribution, $\hat{q}_{enc}(\vec{h} | X_i, \vec{a}_i)$, with mean, $\hat{\mu}_i(X_i, \vec{a}_i)$, and a diagonal covariance, $\hat{\sigma}_i(X_i, \vec{a}_i)$. The neural network, $f_E(X_i, \vec{a}_i, \vec{\beta}_E)$, estimates the distributional parameters, $\hat{\mu}_i(X_i, \vec{a}_i)$ and $\hat{\sigma}_i(X_i, \vec{a}_i)$, for the image $X_i$ and attributes $\vec{a}_i$.

The encoder loss function includes two terms. The first loss term rewards the encoder for estimating $\hat{q}(\cdot\, | X_i, \vec{a}_i)$ that is close to the prior. This term acts to regularize the embedding and prevents the encoder from "cheating" by assigning each image to a unique sub-region of the embedding space. If the encoder were to "cheat," it would effectively memorize training data at the expense of generalizable

performance. We use a standard normal distribution as a prior and measure the distance between distributions by the Kullback-Leibler (KL) divergence (Kingma and Welling 2013).

In practical applications, product attributes $\vec{a}_i$ often contain missing values. The second term in the encoder imputes the missing values by estimating a multinomial classifier. The multinomial classifier is consistent with assuming a Dirichlet distribution of product attributes $\vec{a}_i$ in the probabilistic formulation (see Appendix A.2). For image $X_i$, the encoder neural network produces a probability, $\hat{a}_{ic\ell}$, that image $X_i$ has attribute values $a_{icl}$ for each level, $\ell$, of each attribute, $c$. For example, if Cadillac is a level ($\ell$) of the brand attribute ($c$), then $\hat{a}_{ic\ell}$, is the probability that image $X_i$ is a Cadillac.

Putting these ideas together, the encoder loss function becomes:

(4)
$$\mathcal{L}_{enc}(\vec{\beta}_E) = \sum_{i \in rated, unrated} \left\{ -D_{KL}\left( q_{enc}(\vec{h}|X_i, \vec{a}_i) || \mathcal{N}(0, I) \right) + \sum_{c=1}^{C} \sum_{\ell=1}^{\ell_c} a_{ic\ell} \log \hat{a}_{ic\ell} \right\}$$
$$= \sum_{i \in rated, unrated} \left\{ -\sum_{k=1}^{K} \left[ \frac{1}{2}\left( \mu_{ik}^2 + \sigma_{ik}^2 \right) - \log \sigma_{ik} \right] + \sum_{c=1}^{C} \sum_{\ell=1}^{\ell_c} a_{ic\ell} \log \hat{a}_{ic\ell} \right\}$$

where $D_{KL}$ indicates the KL divergence, $\mathcal{N}(0, I)$ is the standard normal prior, and $k$ indexes the embedding dimensions. The diagonal covariance structure of $\hat{\sigma}_i(X_i, \vec{a}_i)$ and standard normal prior provide a simple representation for the first term in the encoder model (Kingma and Welling 2013).

**4.2. Modification with Adversarial Terms**

Effective machine learning augmentation requires that the model encodes <u>and</u> generates images well. For example, if we are generating luxury crossover utility vehicles (luxury CUVs), the images should look like well-designed luxury CUVs. After extensive experimentation and tuning, we found it necessary to augment the VAE formulation presented in §4.1 using the concept of adversarial training found in generative adversarial networks (GANs).

The basic idea is that we reward the generative model for generating images with embedding distributions similar to the prior, and we reward the encoder for encoding the generated images with distributions distant from the prior. To achieve these joint goals, we implement competing adversarial objectives—a term in the generator is the negative of a term in the encoder, similar to Heljakka, Solin, and Kannala (2019). We train the generator and encoder iteratively, so that the generator and encoder reach a minmax solution to a two-player game. That is, iterative training converges to a fixed point where generated images and actual images are both encoded to the same embedding space (Goodfellow et al. 2014; Ulyanov, Vedaldi, and Lempitsky 2018). Iterative training assures the adversarial terms do not

simply cancel out:

$$\mathcal{L}_{adv}(\vec{\beta}_E) = \sum_{\substack{g \in generated \\ images}} D_{KL}\left(q_{enc}(\vec{h}|X_g, \vec{a}_g)||\mathcal{N}(0,I)\right)$$

(5)

$$= \sum_{\substack{g \in generated \\ images}} \left\{\sum_{k=1}^{K}\left[\frac{1}{2}\left(\mu_{gk}^2 + \sigma_{gk}^2\right) - \log \sigma_{gk}\right]\right\}$$

This approach to adversarial training differs from conventional GANs and VAE-GAN hybrids such as adversarial autoencoders in that we are not learning an implicit generative model (i.e., a likelihood-free generative model specification) by rewarding a "discriminator" to classify images as real or generated. We instead perform adversarial training in the embedding space much like feature matching and perceptual similarity approaches (Larsen, Sønderby, and Winther 2015). Our model is an explicit generative model (i.e., parametric assumptions of the embedding distribution) which ultimately aligns with our managerial use case—a smooth and controllable embedding that allows "creative" exploration by product designers.

### 4.3. Summary

The components of the loss terms described in §4.1 follow the VAE perspective. These components may be viewed as combining semi-supervised VAEs (Kingma et al. 2014) with conditional VAEs (Sohn, Lee, and Yan 2015). We augment the VAE perspective with adversarial methods (similar to GANs) to encourage the generator to produce realistic new images. In contrast to typical VAE-GAN approaches (Larsen, Sønderby, and Winther 2015; Zhao et al. 2016; Berthelot et al. 2017), adversarial autoencoders (Makhzani et al. 2015), and adversarial generative encoders (Heljakka, Solin, and Kannala 2018; Ulyanov, Vedaldi, and Lempitsky 2018), we retain the probabilistic interpretation of the combined model. The proposed approach enables us to sample from the distribution implied by the generator, improves embeddings, and minimizes "posterior collapse" in VAEs.

Table 1 summarizes the loss functions and indicates which images are included in summations. The loss terms are weighted in summation to balance the quality of predictive and generative tasks, and to improve model convergence (§5).

**Table 1.** Predictive, Generative, and Encoding Loss Functions

| Loss Function | Data | Intuition |
|---|---|---|
| **Predictive Model** | | |
| $\lvert y_i - \hat{y}_i \rvert$ | Labeled | MAE term rewards the predictor for predicting ratings. |
| **Generative Model** | | |
| $\frac{1}{3D}\sum_d \lvert x_{id} - \hat{x}_{id} \rvert$ | Labeled & Unlabeled | MAE reconstruction term rewards generator for generating images that are similar to true images. |
| $\frac{1}{D}\sum_d \lvert m_{id} - \hat{m}_{id} \rvert$ | Labeled & Unlabeled | MAE reconstruction term rewards generator for predicting masks that correctly segment the design within the image. |
| $\sum_{k=1}^{K}\left[\frac{1}{2}\left(\mu_{gk}^2 + \sigma_{gk}^2\right) - \log \sigma_{gk}\right]$ | Generated | **Adversarial term** rewards the generator for images with embedding distributions close to the prior. Summed over generated images only ($g$). |
| **Encoding Model (summed over rated, unrated images, or, when indicated, generated images)** | | |
| $\sum_{k=1}^{K}\left[\frac{1}{2}\left(\mu_{ik}^2 + \sigma_{ik}^2\right) - \log \sigma_{ik}\right]$ | Labeled & Unlabeled | KL divergence rewards the encoder for embeddings close to the prior. |
| $-\sum_{c=1}^{C}\sum_{\ell=1}^{\ell_c} a_{ic\ell} \log \hat{a}_{ic\ell}$ | Labeled & Unlabeled | Cross entropy rewards encoder for predicting attributes from images. |
| $-\sum_{k=1}^{K}\left[\frac{1}{2}\left(\mu_{gk}^2 + \sigma_{gk}^2\right) - \log \sigma_{gk}\right]$ | Generated | **Adversarial term** rewards encoder for encoding generated images with distributions far from the prior. Summed over generated images only ($g$). |

## 5. Moving from Theory to Practical Implementation

Our proposed model differs from standard VAE approaches because we include information from product ratings and attributes, and we add masks and adversarial terms. These practical adjustments are necessary to enhance predictive ability and generate realistic new aesthetic designs; however, they create additional technical challenges in model training. We describe a proposed custom deep learning architecture and our approach to training the resulting model. While the architecture and model hyperparameters are specific to our application, the principles behind the modeling and tuning decisions are general. Once trained, our model is rapid and easy to use. We provide a proof-of-concept application with SUVs in §6-7 and discuss an additional proof-of-concept application with dining room chairs in §8.

### 5.1. Deep Learning Architecture

Figure 4 summarizes the deep neural network architectures for the predictive, generative, and encoding models. To simplify the description of the architectures, Figure 4 uses "blocks" of neural network layers, in which each "block" is made up of several neural network layers as described at the bottom of Figure 4. Each neural network layer (e.g., 2D convolution) performs the indicated operations on the outputs from the previous layers.

Starting from the left layer to the right, the 2D convolution layer takes image pixels as input (or the previous layer) and sweeps over patches of pixels using a sliding window of trainable multiplicative weights. This helps the model learn spatial correlations amongst neighboring pixels (or the analogue in higher layers). The spectral normalization layer acts as a regularization technique to control the magnitude of gradients during model training, thereby stabilizing model training (see §5.2). A leaky rectified linear layer acts as a nonlinear activation function to transform values from the previous layer, enabling the neural network to learn complex nonlinear interactions. A residual connection simply adds the original input from the first layer of the block to the now transformed features at the end of the block. In doing so, the intermediary layers learn the residual error from the previous block (Hu, Shen, and Sun 2018). 2D average pooling reduces the dimensionality of the previous layer by down-sampling patches of 2D features to a single value. Squeeze-and-excite explicitly models interdependencies across channels (e.g., RGB channels for the input layer) performing "self-attention". Appendix A.15 provides brief definitions of technical terms used in this paragraph and elsewhere in this paper.

When generating a new image, attributes are run through layers to obtain the $\mu$'s amd $\sigma$'s, which, in turn, generate the predicted encoding, $\vec{h}_r$. Additional layers are not needed in the generator for existing images, hence the dotted box on the left side of the generative model. The predictive model does not have "blocks" but instead fully connected and rectified layers. Lastly, the custom deep learning model requires hyperparameter "tuning," so we hold out data for model selection. We randomly split data into training, validation, and testing sets using a seeded random number generator for reproducibility and statistics. Validation data were used to set model hyperparameters (e.g., learning rates) and monitor training progress for model selection. Testing data were used only for model evaluation.

**Figure 4.** Deep Neural Network Architectures for Predictive, Generative, and Encoding Models

### 5.2. Stabilization and Tuning of Model Training

We train the model using first-order optimizers (See Appendix A.3). Deep learning models are often challenging to train because of the large numbers of parameters to be estimated. This is especially true for the models which include adversarial loss-function terms, as the adversarial components can increase training instability (Gulrajani et al. 2017). Appendix A.7 provides two examples of training instability: gradient explosion and posterior collapse. Gradient explosion results in images unrecognizable as vehicles. Posterior collapse results in images that are all the same.

We apply several techniques to stabilize model training. We avoid unbounded "backpropagated" gradients that lead to catastrophic failures in model training (e.g., outputting images of only white pixels) by stabilizing training with the spectral normalization layer in each block. This helps to regularize the model by dividing the raw output weights of each layer by the largest singular value of the matrix of weights (Miyato et al. 2018). We also enforce both soft and hard constraints on the model architecture. Specifically, we bound the variance of the Gaussian random variables in the KL-divergence terms and use floating point safeguards to ameliorate the potential numerical instability introduced by logarithms and various $L_p$-norms.

We tune model training by scaling the contribution of each loss term in Table 1 with user-defined multiplicative weights. This creates a balance between the seven loss terms given their interdependency during training. These weights stabilize model training and are chosen by monitoring convergence metrics on the training data (e.g., the gradient stability), predictive accuracy on the validation set, and the realism of the generated images during training. For example, we overcame a primary source of training instability which occurred when the KL-divergence of the generated images overwhelmed the KL-divergence of the observed images plus the reconstruction loss. To avoid this problem in the best way, we scaled the KL-divergence loss terms within the range of 1/10 to 1/20 relative to the loss terms for reconstructing images, and annealed these terms from zero at the start of training to a maximum value after model convergence. We further control the difference between the KL-divergence of observed images and the KL-divergence of generated images at each training iteration using a fixed margin (Huang et al. 2018).

We lastly use "progressive training" as proposed by Karras et al. (2017). In this approach, we begin training the model at the lowest image resolution in the data, 4 x 4 pixels, and progressively increase the resolution of input and output images, in stages, until we reach the desired image resolution of 512 x 512 pixels. We denote resolution stages by their height and width, e.g., 4 x 4 pixels or 32 x 32 pixels. As one resolution stage of model training progresses into another, we anneal in the next (larger) image resolution by taking a linear combination of two images, an up-sampled version of the lower resolution image, and the higher resolution image. For each resolution stage, we train the model for multiple iterations, in which we smoothly increase the weight of the higher resolution image in the linear combination. The number of iterations is a hyperparameter controlled using validation data (we trained for 1 million iterations per stage). The advantage of progressive training is that, rather than starting model training from a completely blank slate at full resolution, the model learns information about the aesthetic design at lower resolutions. Incremental learning improves stability of model training and reduces training time. Appendix A.5 provides examples of the generated images at different stages of progressive training.

**5.3. Computational Resources**

Training the model takes roughly 1-2 weeks using a multi-GPU workstation (4 x Tesla V100 or 4 x Quadro RTX 8000), with major computational bottlenecks being both GPU clock speed and GPU memory. Because GPU memory is the main determinant of batch size, we opted for GPUs with either 32GB or 48GB of VRAM. We found that larger batch sizes aided training stability, particularly at the largest progressive training resolutions (e.g., batch size of 48 parallelized over 4 GPUs for resolution of 512 x 512).

We expect training times will decrease with subsequent applications and continual advances in machine learning and computational capabilities. Overall training time depends on a variety of factors including (1) the size of the model itself (the number of trainable parameters and in particular, the dimensionality of the embedding), (2) the number of iterations in progressive training stages, and (3) instabilities that require adjustments to the learning rates and the loss term scaling.

We tuned the model to balance training feasibility and the quality of the output. For example, we used the 512-dimensional embedding space in our automotive proof-of-concept. Fewer dimensions limited the model's ability to encode the information from the images and capture nuances in the aesthetic designs. Larger dimensionality required more training time, without additional benefits to the quality of the predictor or generator. We provide examples of the under- and overparameterized models in Appendix A.8. These tuning decisions might be different for other industry applications.

The monetary costs of the proposed machine learning augmentation are low compared to the multi-million-dollar budgets allocated to improving product aesthetic designs (§1). We trained the model on a workstation that costs less than $10,000. After the model is trained, applying the model does not require machine learning expertise and can be done using standard corporate laptops. The data required for training the model are often routinely available within organizations.

**5.4. Potential Model Extensions**

Many applications of deep learning use a "pretrained model," originally trained on a different task (e.g., object detection), and re-purpose it for the desired task (e.g., aesthetic rating prediction) by re-training it on the desired task's data. Prior marketing science research has applied pretrained models to image data for identifying product returns and brand identity (Dzyabura et al. 2020; Liu, Dzyabura, and Mizik 2020), and to textual data for identifying customer needs and predicting sales conversion (Liu, Lee, and Srinivasan 2019; Timoshenko and Hauser 2019). However, pretrained models generally do not exist for generative models. This is particularly true if the generative model has domain-specific requirements, such as the managerial challenges in §2.2 of limited labeled data and ability to generate new aesthetically-appealing images controllable by designers. The demands of interconnecting attributes, masks, ratings, high-resolution images, and adversarial training for generation required a custom architecture.

Our proposed model architecture supports extensions for applications to different industries and aesthetic design objectives. For simplicity of exposition, we presented a model trained for a single aesthetic rating, however design teams are often interested in multiple metrics for aesthetics. For example, in the automotive industry, firms often consider aesthetic "appeal", "sportiness," and "innovativeness" among other aesthetic attributes and collect such ratings in theme clinics. We can extend our model to multiple aesthetic output measures by leveraging the shared embedding space produced by the encoder and calibrating separate predictor networks for each output measure (see §7.2).

Separability in the use of the predictor, generator, and encoder was motivated by the managerial challenge of fitting machine learning into the traditional design process. Separability is also beneficial in accommodating non-stationarity in customer preferences. For example, aesthetic design of many of the cars popular in 2000s may seem outdated in 2020s. Because firms routinely conduct theme clinics, the predictive model can be regularly updated using more-recent data to account for non-stationarity in preferences. This can be done without retraining the encoder or generator: the updated predictor will prioritize different areas of the product space, while the "definition of the vehicle" in the encoder and generator stays unchanged.

Our model is widely applicable, but not without limit. Consider smartphones. The introduction of iPhone changed the definition of aesthetics from the button-based phone to the touchscreen. Apple defined the look as a "black oily pond." Adjusting for the product definitions requires retraining the entire machine learning augmentation. Fortunately, we can use the lessons learned in this initial application to improve tuning in future applications. We also expect tuning to get easier with further developments in hardware, deep learning frameworks, and transfer learning methods.

## 6. Case Study: Aesthetic Design of U.S. Automotive Market SUV/CUVs

Our machine learning augmentation has two goals: (1) predict consumer evaluations of aesthetics for proposed designs and (2) generate innovative aesthetically-appealing designs that spark creativity among the design team and design management. We calibrate and evaluate the proposed model using unique data provided by our automotive partner.

### 6.1. Data: Images Rated in Theme Clinics

We obtained aesthetic ratings for the unique images of 203 SUVs/CUVs from model years (MY) 2010-2014 tested in our partner's theme clinics. Following established procedures that evolved over decades (and market research in general), the firm used screening questions to target consumers who intended to purchase in the target category in the near future (intenders), who were willing to evaluate aesthetics, and who were screened for sufficient attention and consistency. Respondents were incentive-aligned using standard methods, both fiscally and with knowledge their input would guide future aesthetics (Ding et al. 2011). The details of the screening questions and incentive alignment are proprietary to the firm.

A web-based survey was co-located with the theme clinics. Warm-up questions motivated respondents that their ratings would affect aesthetic design and introduced our partner's previously-calibrated pairwise-semantic-differential rating scale for aesthetic “appeal” (i.e., most unappealing to most appealing). The survey anchored respondents’ ratings by asking each respondent to rate five pre-chosen pairs of images—prechosen in pretests to be most divisive on the pairwise ratings scale. The five image pairs were the same for all respondents, but randomly counterbalanced among respondents.

Each respondent rated ten sequential pages of SUVs/CUVs of five images per page. Each SUV/CUV image was presented from the side viewpoint. To test respondent consistency, the 2nd and 8th page and the 3rd and 9th page contained the same images randomly ordered. After extensive pretesting, the survey was implemented by our automotive partner. To maintain consistency among images and mitigate image-color biases, all images were reduced to greyscale and shown with a side viewpoint. Appendix A.6 provides an example rating page. While we cannot rule out a halo effect, it was unlikely—respondents rated many images and were unlikely to know the marketplace success of each vehicle.

To maintain data quality prior to any further analysis, we eliminated respondents who were judged to be inconsistent based on Krippendorff’s $\alpha$ where $\alpha = 1$ – (observed disagreement among like images) / (expected disagreement due to chance). Krippendorff’s $\alpha$ is a generalization of other interrater reliability measures such as Fleiss’ $\kappa$ and Cohen’s $\kappa$ (Krippendorff 2011). The cutoff was $\alpha = 0.75$, which eliminated 38 respondents (21%). This percentage is consistent with those reported in the literature on eliminating inattentive respondents (e.g., Oppenheimer, Meyvis, Davidenko 2009; Hauser and Schwarz 2015). This literature establishes that such elimination procedures result in higher-quality data that is not biased by elimination. Ratings (7,308) from consistent users were aggregated to a mean value for each of the 203 unique SUVs in MY2010-14. We chose to focus on the side viewpoint in this work as a proof-of-concept. Respondents rated vehicles from the side viewpoint, and the same (mean-value for a vehicle) was assigned to all viewpoints +/- 20 azimuth degrees from the sideview in the training data to increase sample size. We evaluate the prediction methods using only the single side viewpoint of labeled images.[3]

In our analysis, we randomly split the rated data into training, validation, and test data at a ratio of 50%:25%:25%. We used three random splits of the data to allow calibrated standard deviations of the predictive results. Each random split was stratified by considering the make and model of the vehicles. For example, for a given random split, all Jeep Wranglers were in either the training, validation, or test data. This stratified splitting avoids data leakage across year-make-model combinations of vehicles in the data.[4]

---

[3] Appendix A.13 provides an analysis where a model is trained using side views only. Augmenting training data with multiple viewpoints (+/- 20 azimuth degrees) improves the predictive performance.

[4] Due to the stratified sampling, the data split ratios (50%:25%:25%) are approximate.

### 6.2. Data: Unrated Full-Color Images

We obtained access to industry-standard high-quality images available from aggregators. These images are often used by automotive firms in their marketing communications. The typical "rental" cost is about $50,000 per month. We obtained 180,000 unlabeled images of 4,984 unique vehicles across several segments (e.g., sedans, trucks, SUVs). All images were rescaled to 512 x 512 pixel resolution. We used conventional computer vision tools to obtain masks (GrabCut), car color, and viewpoint. Unlabeled images included product attributes such as brand, body type, model, and model year. We describe the available attributes in Appendix A.9.

The unlabeled images were randomly split according by the same process as the labeled images. Unique vehicles held out in the validation and test sets of the labeled image data were held out from the unlabeled image data, thereby ensuring the model never had access to these vehicle images during training. The vast majority of the unlabeled images remained in the training set, because the number of unique vehicles in the unlabeled images dwarf the number of unique SUV/CUVs in the labeled images.

## 7. Evaluation of the Machine Learning Augmentation

We evaluate the ability of the predictor to predict the aesthetic ratings of the held-out vehicles. We evaluate the generator on the face validity of the generated images, the ability to generate images with high aesthetic ratings, the ability to motivate the descriptive insights for new designs, and the ability to generate images comparable to model-year 2020 vehicles that were introduced to the market 5-6 years after model years 2010-2014. Note that our data contains images with different viewpoints, so the *generator* can create new designs with different rotational angles. However, the aesthetic ratings are only available to greyscale images from the side viewpoint (§6.1). We train and evaluate the *predictor* using these data.

### 7.1. Predictive Ability

Figure 5 illustrates predictions for eight SUVs/CUVs; we report the mean absolute error (MAE) for predicted-versus-actual ratings on random splits of the held-out data in Table 2. Our model yields a MAE of 0.350 of a scale point, an improvement of 43.5% over the naïve (uniform) baseline. To put the MAE of the proposed model in perspective, we compare its predictive ability to a series of benchmark models that vary from naïve to sophisticated. We used the same training and validation data to develop the benchmarks and optimized their hyperparameters to provide meaningful comparison.

**Figure 5.** Examples of Predictive Accuracy of Machine Learning Augmentation for Aesthetic Appeal Score

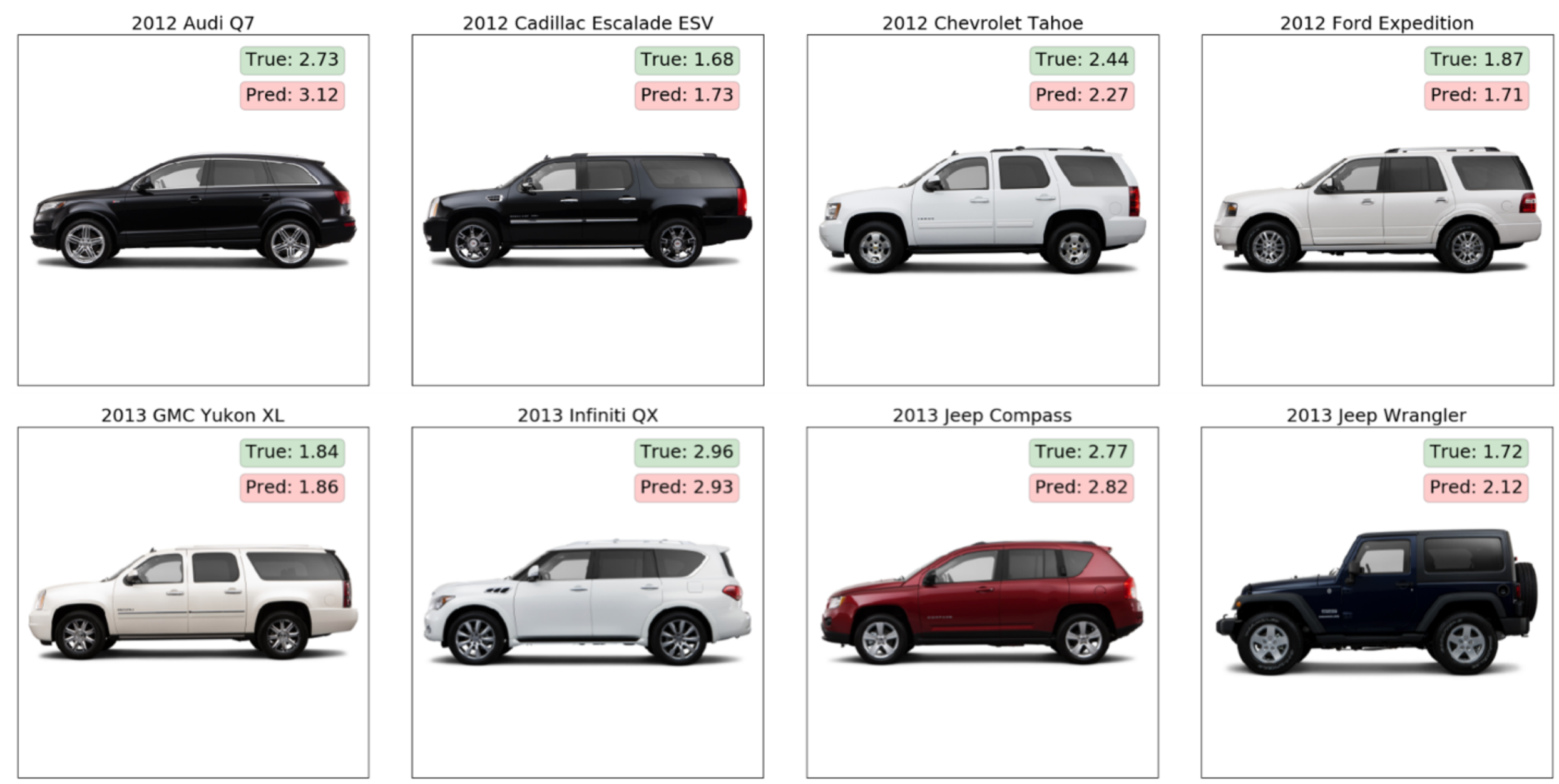

**Uniform baseline**. The most naïve baseline is that respondents select the scale midpoint. This baseline represents zero information. A less naïve baseline uses global information from the training samples to calculate the median rating. To be conservative, we use this less-naïve uniform baseline.

**Sophisticated benchmark 1: Random forest and computer vision features.** Computer vision and machine learning have a long history of processing high-dimensional image and video data for object detection and image segmentation. Conventional approaches reduce high-dimensional visual data to a small set of "hand-engineered" features, which are then input to machine learning methods such as support vector machines.

Our benchmark uses three types of hand-engineered features from computer vision: (1) Histograms of oriented gradients (HOG) features encode edge and shape information. HOG features divide the image into a grid of image patches, calculate the gradients of each patch, and bin these gradients into a histogram. Edge orientation and shape intensity are contained in the gradients' direction and magnitude values. (2) A downscaled version of the image itself (e.g., 512 x 512 to 32 x 32). And (3) histograms of color values for each red, green, and blue image channel. These features are used in a random forest with 100 trees. We present a random forest because it performed best when tested against other common machine learning approaches: support vector machines, Gaussian process regression, and L1/L2-regularized linear regression.

**Sophisticated benchmark 2: Pretrained deep learning model.** Many researchers use "pretrained" open-source neural networks trained for one prediction task and repurposed for another prediction task.

As a sophisticated benchmark, we used the pretrained VGG16 deep learning model trained on the ImageNet database. This benchmark outperformed other common pretrained models (e.g., ResNet50, InceptionV3, YOLOv5) for our prediction task, a finding consistent with the machine learning literature. The VGG16 model is a pyramid of sixteen stacked layers (13 convolutional and 3 fully connected) that sequentially reduce images in size until they are classified in the last layer.

The initial layers of VGG16 transform pixels to edges and lines found in visual images. For our benchmark, we maintain the initial "pretrained" layers and replace the last classification layer with two batch-normalized rectified-linear layers followed by a regression layer. This architecture was chosen using validation data and mirrors the predictive model in our proposed approach. We train the model in two steps. We first freeze the pretrained layers and train only the new layers, then we "finetune" the entire neural network by training all layers. The two-step procedure improves the benchmark's prediction.[5]

**Table 2.** Predictive Test of Machine Learning Augmentation vs Baselines and Benchmarks

| Prediction Model | Mean Absolute Error (std. dev.) | Improvement |
|---|---|---|
| Baseline: Median Rating in Training Images (Constant Rating) | 0.620 (0.043) | 0.0 % |
| Benchmark: Computer Vision Features and Random Forest (Conventional Machine Learning) | 0.446 (0.047) | 28.1 % |
| Benchmark: VGG16 with Fine-Tuned Final Layers (Pretrained Deep Learning) | 0.405 (0.039) | 34.7% |
| Proposed Machine Learning Augmentation (Custom Deep Learning) | 0.350 (0.043) | 43.5 % |

**Results**. Table 2 compares the predictive performance of our proposed machine learning approach to a naïve baseline (the median rating in the training data) and the sophisticated benchmarks for predicting the aesthetic appeal score. The proposed machine learning augmentation outperforms the other methods.[6] Contextualizing this improvement is important. For some applications, particularly those that predict without generation, pretrained networks may be enough. For many firms, however, product design is a multimillion-dollar investment decision, and even a small improvement in precision is valuable as is integration with generation. Our research partner judges that our model's predictions are sufficiently accurate to provide a viable alternative for initial screening prior to formal theme clinics.

[5] For completeness, we used two pretrained deep learning baselines: one without attributes (Table 2) and another with attributes (Appendix A.10). The results are consistent; attributes do not improve predictive performance.
[6] We get the same relative insights if the data are ranks not ratings.

We can evaluate how sensitive the predictive performance of the proposed model and benchmarks to the amount of the training data in our context. We train the models with random subsamples of the labeled and random subsamples of the unlabeled data, and then report the out-of-sample MAE in Appendix A.11 and A.12. The number of labeled images is relatively sparse compared to the unlabeled images. As we reduce the size of the labeled data, the predictive performances of both the proposed model and the benchmark pretrained model deteriorate, and the MAE of the proposed model with 10% labeled data is below 0.5 points on the 5-point scale. The unlabeled images are plentiful. The predictive performance remains similar to the full-data benchmarks down to 10% of the unlabeled data. We notice that while predictive performance is relatively insensitive to the number of unlabeled images at this scale, more images enhance the quality and ease of training of the generator (see Appendix A12).

### 7.2. Generative Capability

By its very nature, the quality of a generated image, and its usefulness to managers and designers, is subjective. Full evaluation is likely to take many years as machine learning augmentation becomes part of an ongoing design process, as new vehicles are launched to the market using the augmented design process, and as we observe market acceptance. An A/B experiment through to market launch is not feasible given organizational constraints and the billion-dollar costs of launching redesigned A vs B vehicles. The best we might hope for is a natural experiment where one suborganization adopts the model and another does not (e.g., Griffin and Hauser 1993 for House-of-Quality adoption at an automotive manufacturer). Even then, only organizational judgments were feasible. At this time, we triangulate the value of the generated images in four ways: face validity, consumer evaluations of generated designs, managerial judgment, and the ability to generate images that are close to innovative vehicles that have been launched after the time during which the training data were obtained.

**Face validity: Controllably generating images**. Our first test is whether or not the proposed approach can create realistic images controllable by attributes (e.g., body type). We begin by sampling points in the embedding space, conditioned on desired attributes, then move smoothly around that space. We use spherical linear interpolation to sample new points. For each point in the embedding space, we generate a high-dimensional image. The images are realistic and can be morphed in a controllable manner that mimics the manner in which design teams create designs. So that the reader may judge, we provide examples in Figure 6 and demonstration videos of SUV/CUV morphing at https://vimeo.com/497011714/ We demonstrate controllability by morphing other body types at https://vimeo.com/334094197.

**Figure 6.** Examples of Generated Designs. 1st Row: SUV/CUV. 2nd Row: Sedan to Truck. 3rd Row: Rotation

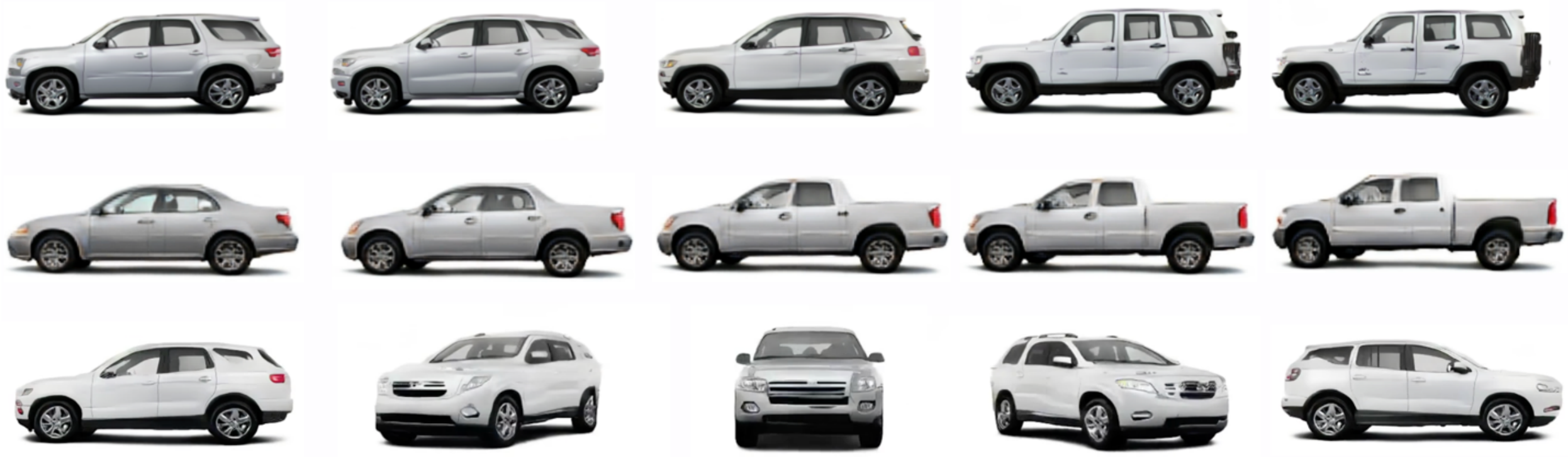

**Generating appealing images: Consumer evaluations**. To test the ability of the model to produce aesthetically-appealing images, we generated 50 targeted images: 25 were predicted by the predictor to be rated highly and 25 were predicted to be rated poorly. Figure 7 provides examples of generated images of each type. For consistency with the training data available to the *predictor*, we generated each image to be a light gray SUV/CUV from the side view. The generated designs were created by using spherical interpolation between existing designs to ensure plausibility to respondents and to mitigate biases (Lopez, Miller, and Tucker 2019). Following §6.1, we used respondents from a professional Internet panel (ProdegeMR, at $4 per respondent) to evaluate the aesthetic appeal over randomly-selected pairs of the generated designs. Following industry standards (also used by our industry partner), we screened respondents to be SUV/CUV "intenders."

We pretested the survey carefully. The initial sample was 358 respondents. Following suggested practice and prior to any analysis of the data, we used instructional manipulation checks (IMCs) to eliminate 116 inattentive and/or "professional" respondents (Oppenheimer, Meyvis, and Davidenko 2009). In particular, respondents were eliminated if they were not SUV intenders, answered too quickly or too slowly, responded with "straight-line" patterns, or failed "trap" questions that tested for attention. Our elimination rate is typical of industry and academic experience—see review in Morren and Paas (2019). IMCs increase the reliability of survey data and encourage respondents to think hard (Hauser and Schwarz 2015; Oppenheimer et al. 2009). The final screen to 181 respondents eliminated an additional 61 respondents who were inconsistent in answering repeated binary choice questions. (We built the repeated questions into the survey before the survey was fielded.)

The consumer evaluations suggest that respondents judged as aesthetically appealing images that the predictor forecast to be aesthetically appealing, and judged as aesthetically unappealing images that the predictor forecast to be aesthetically unappealing. Specifically, the predictor and consumers agreed 74.0% of the time.

**Other aesthetic metrics: Innovativeness**. Firms are often interested in multiple aesthetic metrics, and our model can be easily recalibrated for different metrics. Our industry partner provided ratings for aesthetic "innovativeness" over the same unique 203 SUV/CUVs in the labeled data as for aesthetic "appeal". We recalibrated the model to predict aesthetic innovativeness by finetuning a previously-trained model of aesthetic appeal using the aesthetic ratings for "innovativeness" (see Appendix A.14 for details). We then sampled SUV/CUV designs from the embedding space. Figure 7 illustrates vehicles with predicted high- and low-aesthetic-appeal and with predicted high- and low-aesthetic-innovativeness.

**Figure 7**. Example of Generated Designs for Consumer Evaluation and Augmenting Managers

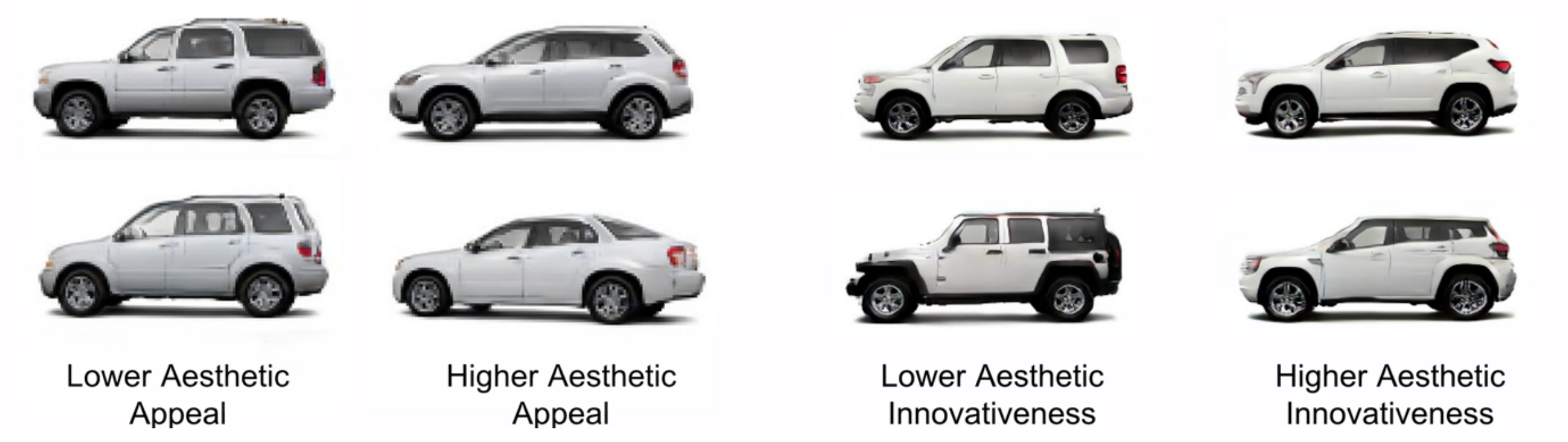

Figure 7 highlights that aesthetically appealing and aesthetically innovative designs can take very different aesthetic forms. One cannot exhaustively describe the differences between the designs in Figure 7 with a small set of attributes. Generated images are instead intended to visually showcase these differences to guide management and spark creativity amongst designers.

**Augmenting managers: Can the model guide design exploration?** Design managers balance current consumer preferences with designers' creative visions of the future. Firms aim to strike a delicate balance between a vehicle's aesthetic appeal and aesthetic innovativeness: a notion supported by academic literature—too much aesthetic innovativeness and the product is overly Avant-garde and unlikely to be appealing to a large market; too little innovativeness and the product quickly becomes stale with lacking competitive advantage (Landwehr, Labroo, and Herrmann 2011; Toubia and Netzer 2017).

We showed the images in Figure 7 to senior managers at our automotive partner who were responsible for evaluating aesthetic design.[7] These managers immediately recognized differences between the generated designs and attempted to identify design factors associated with aesthetic innovativeness for new vehicles (e.g., neutral “rake” with negatively sloping roof). The images inspired the managers to consider other design features for investigation such as the “front overhang” and “hood slope.” These managers stated further that deep generative models are cost effective, inform and augment designer intuition, and may offset any (human) biases in the design generation and selection process. Although this evidence is anecdotal, the images seemed to be extremely valuable to guide exploration of the design space by experienced senior managers.

[7] Andrew Norton is an Executive Director of Global Market Research, Volume Forecasting, and Competitive Intelligence at General Motors. Jeff Hartley is an Adjunct Associate Professor of Integrative Systems and Design at University of Michigan. He worked as a Technical Director at General Motors for over 30 years.

**Anticipating successful designs.** One of the first questions practicing designers ask is whether the model can generate “creative” designs. Our model was only trained on data from MY2010-2014. Many new aesthetic designs have since been introduced to the market than are not in our training data. As a minimal test of the ability to produce known creative designs, we examine whether the generator could have produced images that are similar to since-introduced MY2020 vehicles.

Figure 8 compares four SUV/CUV designs from the MY2010-14-trained generator to four new SUV/CUV designs from MY2020. While not identical, the generated images evoke holistic aesthetics of the recently-introduced vehicles. At a more-detailed level, the comparative designs are similar on common measures of vehicle aesthetics such as proportion, surface, and detail (PSD). For example, the second column contains a generated design with very high and positively-angled “beltline” coupled with a dramatically downward-swooping “greenhouse,” a design later introduced in the Mercedes GLE. Because new aesthetically-appealing PSDs are of particular interest in designers’ creative visions, it is encouraging that the generator discovers designs that have PSDs comparable to new production vehicles introduced successfully to the market six years after the time frame from which the training, validation, and test data were drawn.

**Figure 8.** Examples of Generated Designs (MY2010-14 Data) and Actual Production Designs (MY2020)

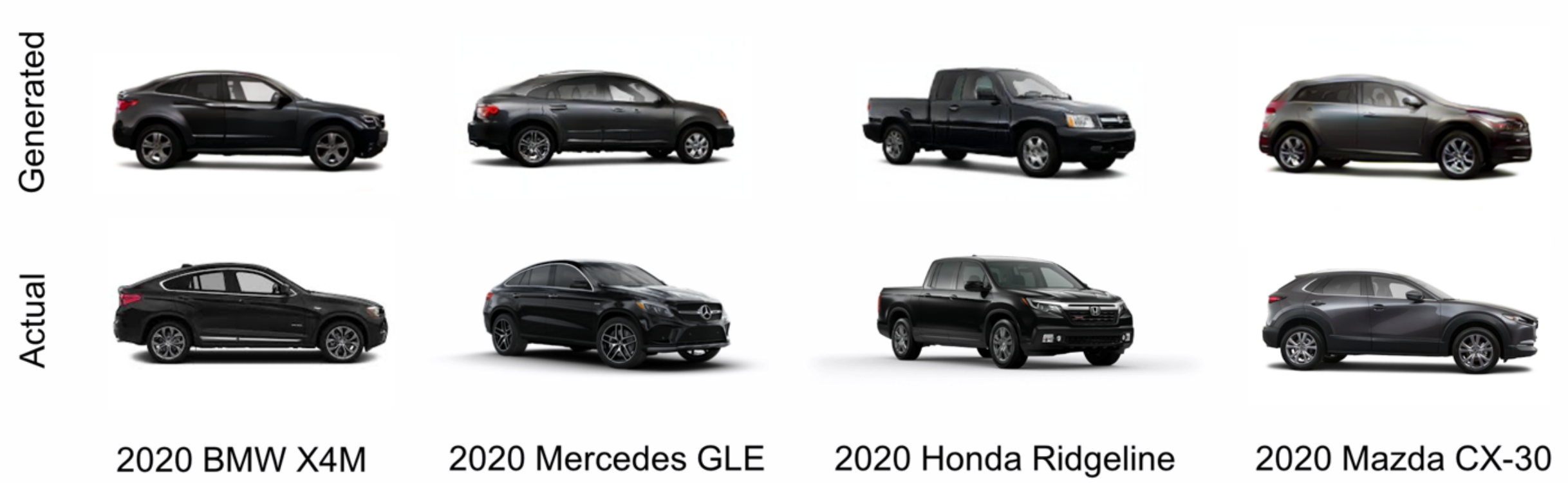

## 8. Applications in Other Categories: Dining Room Chair Example

We engineered a machine learning augmentation that would apply generally across product categories. We chose the automobile industry for our initial proof-of-concept application because product aesthetics are particularly valuable to the automobile industry. Our partner provided us with a unique opportunity to understand organizational needs and access to the same proprietary data used routinely by human designers. In this section, we explore an additional application using publicly available images of dining-room chairs. We collected aesthetic ratings from an online panel of responders using the same procedure described in §7.2, and trained and tuned the model as described in §4 and §5. For replication, we provide our codebase as opensource.

### 8.1. Dining-Room-Chair Images

The images for our second application come from an opensource dataset of chair images provided by Aubry et al. (2014). These images were created by rotating 3D computer-aided-design (CAD) drawings of chairs, and taking 2D image snapshots across 62 angular viewpoints for each chair. The market for chairs, like the automotive market, is segmented; predictions make the most sense within a segment. We chose the dining-room-chair segment which had 700 unique dining room chairs—one of the largest segments in the sample—for a total of 43,400 images (62 times 700) across all viewpoints. We preprocessed the images to grayscale and downscaled to a variety of resolutions (8 x 8, 16 x 16, …, 128 x 128) for progressive training. We reparametrized the image viewpoints to consistent angular coordinates.[8]

To implement the augmentation model consistently with the automotive application, we obtained aesthetic ratings (labels) for 200 of the 700 dining room chairs. Two hundred labeled images are comparable in number to the 203 unique labeled automobiles (SUV/CUVs) (§6.1). Based on small-sample qualitative research and an initial pilot test with 101 Amazon Mechanical Turk respondents, we selected a 5-point semantic aesthetic scale from "Very Traditional to Very Modern" as most descriptive, most consistent, and least ambiguous aesthetic dimension.[9] For this aesthetic scale, the survey respondents provided consistent ratings for each chair design. The average rating varied across the chairs. We sourced 510 new respondents from the same professional panel used for the automotive data (ProdegeMR). Instructional manipulation checks filtered our sample to 348 attentive respondents.

---

[8] Because the dining-room-chair images are substantially fewer and with lower resolution than the automotive images, our second application serves to test how our models perform with smaller and lower quality data.

[9] We selected "traditional vs. modern" over "obtrusive vs. prominent" and "typical vs. unique" based on respondent consistency in the MTurk pretest. In open-ended interviews, consumers found these three dimensions to be relevant, easy-to-evaluate, well-defined, and unambiguous. Details available from the authors.

### 8.2. Model Training and Predictive Test

We trained the proposed model for the new category up to the highest resolution consistent with the opensource images, 128 x 128. We used the same procedures as in the automotive application. For the predictive tests, we use an equivalent baseline and the same computer-vision and pretrained deep learning models. Table 3 displays the predictive tests for the dining-room-chair images. The custom deep learning model outperforms both conventional machine learning and pretrained deep learning. Interestingly, conventional machine learning outperforms pretrained deep learning. Conventional machine learning, tuned to predictive ability, is almost as good as that of the custom deep learning model, which must also predict and generate. The lower quality of the unlabeled dining-room chair images likely puts an upper bound on the predictive ability of any machine learning model, conventional or custom deep learning. At minimum, the predictive test for dining-room-chair images suggests that the custom deep learning model can predict well even when trained on lower-quality images.

**Table 3.** Predictive Test for Dining-Room Chairs

| Prediction Model | Mean Absolute Error (std. dev.) | Improvement |
|---|---|---|
| Baseline: Median Rating in Training Images (Constant Rating) | 0.480 (0.011) | 0.0 % |
| Benchmark: Computer Vision Features and Random Forest (Conventional Machine Learning) | 0.430 (0.016) | 10.4 % |
| Benchmark: VGG16 with Fine-Tuned Final Layers (Pretrained Deep Learning) | 0.434 (0.036) | 9.6 % |
| Proposed Machine Learning Augmentation (Custom Deep Learning) | 0.423 (0.013) | 11.9 % |

### 8.3. Illustration of the Generator

The dining-room-chair dataset used in our application is widely accepted in computer science research focused on generative modeling. In the automotive proof-of-concept (§6-7), we were fortunate to have access to senior managers and to images of successful SUVs launched after the time of data collection. We do not have an industry partner in the furniture category, nor the high-quality proprietary images normally retained by retailers and manufacturers. Nonetheless, we evaluate whether we can generate images that are as dining-room-chair-like as opensource CAD images and literature baselines.

The first row of Figure 9 is a sample of the opensource dining-room-chair CAD images. The second row illustrates chairs generated by our proposed model. Our *generator* sought to generate new aesthetic designs that were not among the opensource images (i.e., our goal is not reconstruction). The third row illustrates the generated images for three established baselines — deep convolutional inverse graphics network (DC-IGN; Kulkarni et al. 2015), information maximizing GAN (InfoGAN; Chen et al. 2016), and a VAE (Kingma and Welling 2013). The baseline models were trained on the same opensource data that we used (Higgins et al. 2016); we provide two typical examples for each of the three models.

**Figure 9.** Example Opensource and Generated Dining-Room Chairs

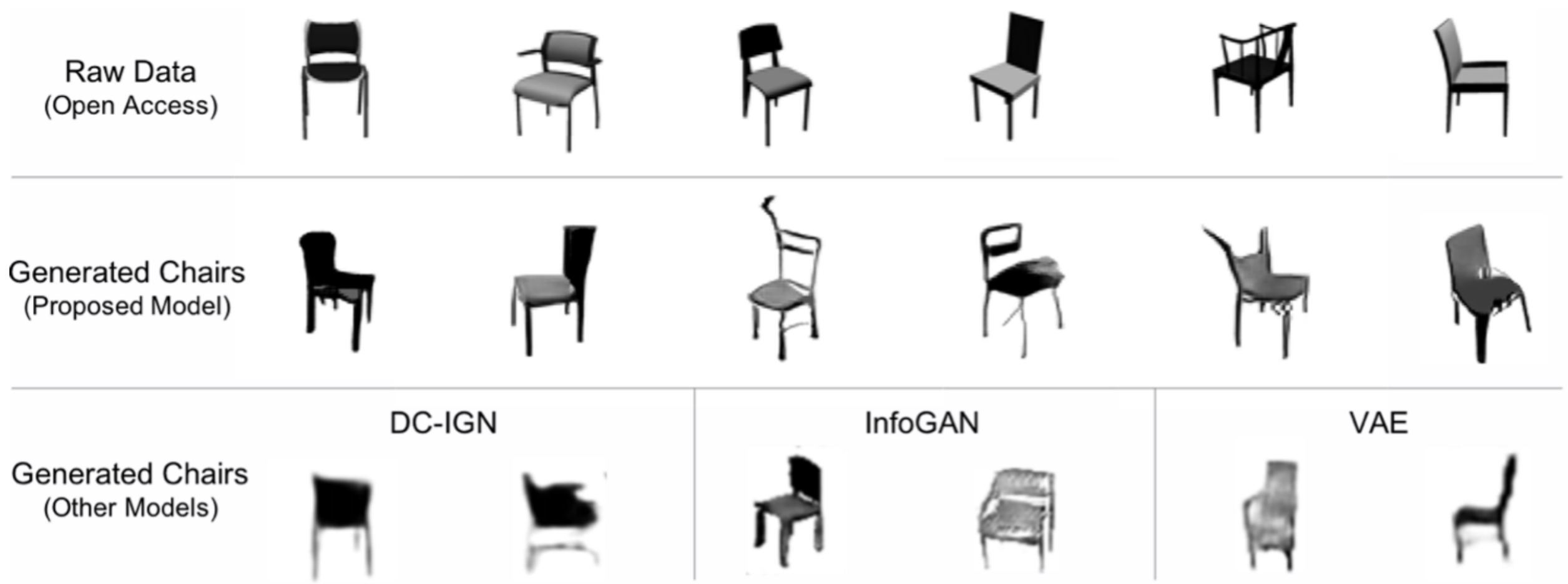

Our model and the established baselines generate images that are dining-room-chair-like. They suggest ideas that a designer might pursue. Our model appears to generate images that are crisper than those generated by DC-IGN and the VAE and are at least as good as those generated by InfoGAN. This observation is in line with our modeling approach that combines the VAE architecture for controllability and adversarial components in training for realism. At minimum, the replication in dining-room chairs suggests our model does as well as established models when trained on sparse CAD images.

The generated dining-room-chair images are not as smooth and complete as the automotive images, likely because of the lower quality of the dining-room-chair image data. Access to sufficiently large samples of high-quality images might be an important requirement for the industry adoption of the generative models for product aesthetic design. Fortunately, many firms routinely generate many high-quality-image aesthetic designs in the normal course of business (See §2.1 and 8.4).

**8.4. Summary of the Replicability Test**

The dining-room-chair application demonstrates that our proposed integrated deep learning model is tractable outside of the automotive industry. We find the dining-room-chair results qualitatively mirror those from the automotive case study. Quantitative differences are likely due to the lower realism of dining-room chair images (2D images from 3D CAD renderings). Conditional on the available data, the predictive and the generative abilities in the replication study are comparable to the state-of-the-art baselines.

The number and quality of unlabeled images likely affects predictive and generative performance. We believe many product categories are amenable to our work. For example, there are 10,000 home-furnishing items in the IKEA catalog. Dyson has several hundred sketches per new product across a large number of products. There are 700,000 SKUs in the product line of a large fashion retailer. Even within apparel segments, there should be sufficiently many high-quality unlabeled images to train the model well.

## 9. Conclusion

### 9.1. Discussion and Summary

Deep learning methods are beginning to affect all aspects of marketing science, sometimes with methods customized to the challenge, sometimes with tuned pretrained models. Many of these methods rely on hard-to-quantify unstructured data such as natural language or images. We focus on consumers' aesthetic judgments of images by using state-of-the-art machine learning to augment human design decisions. The augmentation comes in two forms. First, we predict consumer evaluations of new potential aesthetic designs from images. Second, we controllably generate new images to enhance creative design.

Our machine learning approach combines many concepts with an overall goal of aligning with actual aesthetic design processes used at firms. We developed a version of the semi-supervised VAE model that uses a low-dimensional embedding to "bottleneck" information between a predictive and generative model. Within the VAE framework, we include attributes to carry information about images and include masks to constrain target images to be realistic. We add adversarial training concepts from the GAN literature to improve training of realistic generated images. Finally, we use a variety of engineering ideas (e.g., spectral normalization, progressive training, residual connections, adaptively balanced training losses) to tune the deep learning model so it stably converges during training.

Our proposed augmentation is based on our understanding of how real organizations design product aesthetics. Our model recognizes the delicate and iterative interplay of machine learning and existing human workflows, a point stressed heavily in our working interactions with managers and designers. Our model is separable in use, enabling adoption by asynchronous and distributed design and testing teams. We focus on *augmenting* rather than *automating* human expertise and human creativity, and ensuring all models are meaningfully controllable by the respective teams within the firm.

Our model addresses the practical challenge of using a relatively limited amount of costly-to-obtain labeled images. If we were to train a deep learning model on labeled images alone, the embeddings and generative capability would be weak at best. We overcome the challenge with semi-supervised learning that combines expensive rated "thin data," with less expensive and significantly larger "big data." High-quality unlabeled images are not inexpensive and often protected by copyright, but they are often available to the firm.

We demonstrate that the predictor predicts image ratings better, sometimes substantially better, than strong, tuned machine learning benchmarks such as conventional computer-vision methods and pretrained deep learning models. We demonstrate that the generator (1) generates face-valid images, (2) that consumers evaluate as aesthetically appealing images created to be aesthetically appealing, (3) that generated images anticipate designs that were introduced to the marketplace five-six years later, (4) that the model can be tuned to process other aesthetic scales and (5) can be applied to non-automotive product categories (§8). Anecdotally, the automotive generator is viewed by senior design and marketing managers as valuable and worth further investment.

**9.2. Limitations and Further Research**

The SUV/CUV application is a proof-of-concept developed to augment aesthetic design teams. New vehicles take many years and $1-3 billion in investment (Blonigen, Knittel, and Soderbery 2013). Over time we will learn whether machine learning augmentation has documented monetary benefits beyond the qualitative benefits illustrated in this paper. Directly assessing the financial value of aesthetics is challenging given its interrelatedness of new aesthetic designs with confounding factors such as functional attributes, brand identity, marketing, pricing, and aesthetic trends (Person, Snelders, and Schoormans 2016). Recent promising work into disentangling these factors includes those that explicitly control covariation in functional and form attributes (Higgins et al. 2018; Kang et al. 2016; Zhang et al. 2019), as well as those that temporally model aesthetic trends (Yoganarasimhan 2017). For now, our work relies on predictive statistics, generative illustrations, and managerial judgment for validation.

**Image quality.** Professional-level images enhance quality by controlling for variables that inadvertently affect aesthetic perceptions. Variables include f-stop (e.g., fishbowl, telescopic), zoom level, azimuthal capture angle, chromatic aberrations, lighting (saturation and hue), day vs. night, masking controls, background images, occlusions in foreground images, and visual noise.

Higher resolutions help the model identify aesthetic dimensions not available at lower resolutions and account for how those dimensions combine holistically. This was evidenced in the automotive application when training the model at lower resolutions during progressive training, as well as in the chair application, in which the data was of limited resolution and realism (see §8.4). Given these benefits, it is not surprising that human designers prefer to work with higher-quality (higher resolution) images. However, tuning the machine learning models for higher resolutions requires more training data, as the models need to encode and generate more complex aesthetics. Future research could enhance quality by developing approaches for modeling and evaluating holistic 3D designs (Wu et al. 2016).

**Data needs**. Product categories vary in how challenging they are to model. For example, the aesthetics for dining-room chairs are likely simpler to model than human faces, while human faces are likely simpler to model than automobiles (Karras et al. 2017). While our experience with two proof-of-concept categories suggests it might be sufficient to have unlabeled data in the tens of thousands and labeled data in the hundreds, more applications will pin down data needs. Of course, the more data the better, and advances in machine learning research can further reduce the data needs.

**Scale**. While our model was initially engineered to be effective for automotive vehicles, the principles and modeling decisions generalize to other marketing/aesthetic design applications. We used 180,000 unlabeled images in training—a typical scale for automotive aesthetic design applications. Were the model to be scaled to millions or billions of images, we would likely need to rely on distributed computing and different neural network architectures.

**Technical issues**. Many of the technical challenges came from combining VAE and GAN concepts. When "realistic" generation alone is the primary goal, pure GANs often outperform VAEs and flow-based approaches (Kang et al. 2017; Pan et al. 2017; Sbai et al. 2019). But GANs often lack the latent-space embedding structure needed for predictive modeling and controllable "creative" generation. Further development of deep generative modeling frameworks, improved methods for systematic model tuning, and further experience in other applications and product categories will simplify the calibration of augmentation models (see §5.2).

**Further work with designers**. A natural next step is to assess the degree to which the proposed approach augments designer creativity. While perhaps less straightforward to measure, similar questions have seen recent marketing interest in applications such as idea generation (Toubia and Netzer 2017) and branding and logo generation (Dew, Ansari, and Toubia 2022). Machine learning methods that promote "diversity" of generated designs, for example, augment designer creativity via larger search of the space of designs (Nobari, Chen, and Ahmed, 2021). Likewise, methods such as "disentangled" representation learning may offer designers and managers opportunities to identify new aesthetic attributes (Higgins et al. 2018). As we continue to be guided by real design needs and managerial problems, ongoing advances in machine learning bodes well for a future of augmenting human intelligence with machine intelligence.

**Appendices**

**A.1. Summary of Notation**

| | |
|---|---|
| $X_i$ | product image $i$; contains $D$ pixels $x_{id}$ |
| $M_i$ | product mask; contains $D$ binary values $m_{id}$ |
| $y_i$ | aesthetic rating |
| $\vec{a}_i$ | product attributes vector (if known) |
| $\hat{X}_i$ | product image (generated); contains $D$ pixels $\hat{x}_{id}$ |
| $\hat{M}_i$ | product mask (generated); contains $D$ binary values $\hat{m}_{id}$ |
| $\hat{y}_i$ | aesthetic rating (predicted) |
| $\hat{a}_i$ | product attributes vector (predicted) |
| $\vec{\mu}_i, \vec{\sigma}_i$ | product embedding distribution parameters, $q_{enc}(\vec{h}\vert X_i, \vec{a}_i)$ |
| $\vec{h}_i$ | product embedding vector ($K$-dimensional) |
| $\vec{\beta}_E, \vec{\beta}_P, \vec{\beta}_G$ | estimated parameters of the encoder, predictor, and generator |

**Appendix A.2. Probabilistic Formulation and Loss Function Separability**

To ease notation, we temporarily write all parameters and likelihoods for a single datum $i$. We seek a joint distribution, $p(y_i, X_i|\vec{a}_i, \vec{\beta})$, for the ratings and images conditioned on the design attributes $\vec{a}_i$ and the parameters $\vec{\beta}$. The joint distribution can be decomposed into a predictive model and a generative model by the laws of conditional probability:

(A1)
$$p(y_i, X_i|\vec{a}_i, \vec{\beta}) = p_{pred}(y_i \mid X_i, \vec{a}_i, \vec{\beta}) p_{gen}(X_i|\vec{a}_i, \vec{\beta})$$

Representing and estimating the two conditional distributions is not feasible when product images are high dimensional. To address high dimensionality, we approximate the true joint distribution using *embeddings*. In our case, the embedding compresses information from high-dimensional images, aesthetic ratings, and product attributes to enable tractable predictive and generative models.

We estimate an embedding posterior distribution, $q_{enc}(\vec{h}_i|X_i, \vec{a}_i, \vec{\beta})$, for each product design $i$ such that $\vec{h}_i$ has substantially fewer dimensions (e.g., 512) than $X_i$ (e.g., 786,432), yet retains most of the information contained in the images, ratings, and product attributes. To infer embeddings, we use variational Bayes methods to approximate the true joint log-likelihood, $\log p(y_i, X_i|\vec{a}_i, \vec{\beta})$, with an approximate log-likelihood, $\ell^i_{approx}(\vec{\beta})$, dependent on the embeddings, $\vec{h}_i$ (Blei, Kucukelbir, and McAuliffe 2017; Jordan et al. 1999).

To obtain this approximation, we first condition the true log likelihood, $\log p(y_i, X_i|\vec{a}_i, \vec{\beta})$, on the embeddings via marginalization, which leads to the logarithm of an expectation. We then approximate the logarithm of the expectation by the expectation of the logarithm. By Jensen's Inequality, the approximation is a lower bound to the true log likelihood. Rearranging terms we arrive at the approximate likelihood in Equation 1 in §4. Given an image, $X_i$, its rating, $y_i$, and its attributes, $\vec{a}_i$, we seek to maximize $\ell^i_{approx}(\vec{\beta})$ and thus approximately maximize $\log p(y_i, X_i|\vec{a}_i, \vec{\beta})$. If $D_{KL}(\cdot \,||\, \cdot)$ signifies the Kullback-Leibler (KL) divergence, then:

(A2)
$$\begin{aligned}\ell^i_{approx}(\vec{\beta}) &= \mathrm{E}_{\vec{h}_i}\left[\log p_{pred}\left(y_i|\vec{h}_i, \vec{\beta}\right) + \log p_{gen}\left(X_i|\vec{h}_i, \vec{\beta}\right) - \log q_{enc}\left(\vec{h}_i|X_i, \vec{a}_i, \vec{\beta}\right)\right. \\ &\quad \left. + \log p_{prior}\left(\vec{h}_i|\vec{a}_i\right)\right] \\ &= E_{\vec{h}_i}\left[\log p_{pred}\left(y_i|\vec{h}_i, \vec{\beta}\right) + \log p_{gen}\left(X_i|\vec{h}_i, \vec{\beta}\right)\right] \\ &\quad - D_{KL}\left(q_{enc}\left(\vec{h}_i|X_i, \vec{a}_i, \vec{\beta}\right)||p_{prior}\left(\vec{h}_i|\vec{a}_i\right)\right)\end{aligned}$$

Equation A2 is intuitive. The first term seeks to maximize our ability to predict aesthetic ratings based on the embeddings. The second term seeks to reproduce images based on the embeddings. The last term is the negative of KL divergence from the embedding (distribution) to the prior on the embeddings. For this derivation, we have assumed conditional independence between images, ratings, and attributes given the embedding.

Because the approximate log-likelihood, $\ell^i_{approx}(\vec{\beta})$, is a lower bound to the true log likelihood, $\log p(y_i, X_i|\vec{a}_i, \vec{\beta})$, the approximation is commonly called the "evidence lower bound" (ELBO; Jordan et al. 1999). We derived Equation A1 for every datum $i$, which leads to an overall full data approximate log-likelihood, $\mathcal{L}(\vec{\beta})$, which we maximize over parameters, $\vec{\beta}$, using the observed images and ratings:

(A3)
$$\mathcal{L}(\vec{\beta}) = \sum_i \ell^i_{approx}(\vec{\beta})$$

We use conditional probability to separate Equation A2 into three component models:

(A4)
$$\mathcal{L}(\vec{\beta}) = \mathcal{L}_{pred}\left(\vec{\beta}_P\right) + \mathcal{L}_{gen}\left(\vec{\beta}_G\right) + \mathcal{L}_{enc}\left(\vec{\beta}_E\right)$$
$$\mathcal{L}_{pred}\left(\vec{\beta}_P\right) = \sum_{i \,\in\, rated} E_{\vec{h}_i}[\log p_{pred}(y_i|\vec{h}_i, \vec{\beta}_P)]$$
$$\mathcal{L}_{gen}\left(\vec{\beta}_G\right) = \sum_{i \,\in rated, unrated} E_{\vec{h}_i}[\log p_{gen}(X_i \mid \vec{h}_i, \vec{\beta}_G)]$$

$$\mathcal{L}_{enc}(\vec{\beta}_E) = - \sum_{i \in rated, unrated} \Big\{ D_{KL}\left( q_{enc}(\vec{h}_i | X_i, \vec{a}_i, \vec{\beta}_E) || p_{prior}(\vec{h}_i | \vec{a}_i) \right) + D_{KL}\left( q_{attr}(\vec{\pi}_i | X_i, \vec{\beta}_E) || \, p_{attr}(\vec{\pi}_i | \vec{a}_i) \right) \Big\}$$

In our empirical application, for (relatively) less expensive unlabeled data, we have access to product attributes (e.g., color, brand). Accordingly, the encoder log-likelihood, $\mathcal{L}_{enc}(\vec{\beta}_E)$, differs from that in Equation A2 because we also learn the relationship between images and their attributes. Thus, in addition to the embeddings, $\vec{h}_i$, we use variational inference to encode attribute information with $\vec{\pi}_i$. Following the same reasoning used to derive Equation A2, we obtain the last KL divergence term in the encoder log-likelihood. This term acts as a classifier for attributes, $\vec{a}_i$, when the attributes are known. We use the classifier to predict attributes when they are unknown or ambiguous (e.g., when generating new designs). See also Keng (2017).

We choose probability distributions for the predictive, generative, and encoding models in Equation A4 using the framework of variational autoencoders (VAE; Kingma and Welling 2013). Under the proposed distributional assumptions, the log-likelihood formulation is equivalent to the loss-function formulation labeled as Equation 1 in §4.

**Predictive Model.** For the predictive model, we choose Laplace distributions with means $y_i$ and unit diversity, $p_{pred}(y_i \, | \vec{h}, \vec{\beta}_P) = \frac{1}{2} e^{-\left| y_i - f_P\left(\vec{h}_i, \vec{\beta}_P\right)\right|}$. The Laplace distribution is converted to the $L_1$ norm via the log-likelihood, thus enabling a probabilistic interpretation for the absolute loss in Equation A5. This predictive distribution implies that we minimize the mean absolute error of predicted versus true ratings, where $\hat{y} = f_P(\vec{h}_i, \vec{\beta}_P)$:

(A5)
$$\mathcal{L}_{pred}(\vec{\beta}_P) = \sum_{i \in rated} E_{\vec{h}_i}[\log p_{pred}(y_i | \vec{h}_i, \vec{\beta}_P)] = - \sum_{i \in rated} |y_i - \hat{y}_i|$$

**Generative Model.** We choose the generative model to be a high-dimensional Laplace distribution with means, $X_i$, and unit diversity, $p_{gen}(X_i | \, \vec{h}_i, \, \vec{\beta}_G) \, \propto \, e^{-\left| X_i - f_G\left(\vec{h}_i, \vec{\beta}_G\right)\right|}$. Similarly, we assume a high-dimensional Laplace distribution for masks $M_i$. As with the predictive model, the Laplace distribution is naturally conjugate to the $L_1$ norm enabling a probabilistic interpretation for the absolute loss. This implies the following log-likelihood function for the generative model:

(A6)
$$\mathcal{L}_{gen}(\vec{\beta}_G) = \sum_{i \in rated, unrated} E_{\vec{h}_i}[\log p_{gen}(X_i, M_i | \, \vec{h}_i, \vec{\beta}_G)]$$

$$= - \sum_{i \in rated,unrated} \left\{ \frac{1}{3D} \sum_{d} |x_{id} - \hat{x}_{id}| + \frac{1}{D} \sum_{d} |m_d - \widehat{m}_{id}| \right\}$$

**Encoding Model.** For the embedding variational family, we choose multivariate Gaussian mixture distributions with mixture components depending on product attributes, $\vec{a}_i$, (e.g., 'SUV'). The embedding, $\vec{h}_i$, has a Gaussian mixture marginal distribution, but $\vec{h}_i|\vec{a}_i$ has a single Gaussian conditional distribution given attributes (Dilokthanakul et al. 2016). This expands a representation capacity of our model (Ranganath, Tran, and Blei 2015), without resorting to more complicated autoregressive and flow-based methods (Chen et al. 2016).

We further assume each $K$-dimensional Gaussian has diagonal covariance, thereby factorizing into $K$ conditionally independent Gaussians in which $K$ is the dimensionality of the embeddings. If $k$ indexes the elements of the embedding, then the variational assumption implies that $q_{enc}(\vec{h}_i \mid X_i, \vec{a}_i, \vec{\beta}_E) \propto \prod_{k=1}^{K} \sigma_{ik}^{-1} e^{-\frac{1}{2}(h_{ik}-\mu_{ik})^2 \sigma_{ik}^{-2}}$, where the $\vec{\mu}_i$ and $\vec{\sigma}_i$ are functions of $X_i$, $\vec{a}_i$, and $\vec{\beta}_E$. Following Kingma and Welling (2013), we obtain a simpler representation $D_{KL}(\mathcal{N}(\mu_{ik}, \sigma_{ik}) \mid\mid \mathcal{N}(0,1))$.

For the second divergence term in the encoder log-likelihood, we show below that we may approximate $D_{KL}(q_{attr}(\vec{\pi}_i|X_i, \vec{\beta}_E) || \, p_{attr}(\vec{\pi}_i|\vec{a}_i)) \cong constant - \log q_{attr}(\vec{a}_i|X_i, \vec{\beta}_E)$. This results in the encoder acting as a multinomial classifier, $q_{attr}(\vec{a}_i|X_i, \vec{\beta}_E)$, to predict attributes from product images. Specifically, we have $C$ multinomial distributions, where $C$ is the number of attributes (e.g., brand, body type) and $\ell_c$ is the number of levels of attribute $c$.

The encoder neural net for $\vec{\mu}_i$ and $\vec{\sigma}_i$ thus also produces $\vec{\pi}_i$, from which we draw Dirichlet probabilities, $\hat{a}_i = q_{enc}(\vec{\pi}_i|X_i, \vec{\beta}_E)$, using a soft-max function. We recognize $E_{\vec{\pi}_i}[\log q_{enc}(\vec{\pi}_i|X_i, \vec{\beta}_E)]$ as the cross-entropy for a draw of the attributes, $\vec{a}_i$, from the multinomial probabilities, $\hat{a}_i$. This provides the second term in the loss function below. During training, this term encourages the encoder to learn attributes, while during prediction (when we do not know attributes) this term allows us to estimate unknown product attributes, $\hat{a}_i$, by sampling from the multinomial distribution indexed by $\vec{\pi}_i$. Putting both terms together we obtain:

(A7) $$\mathcal{L}_{enc}(\vec{\beta}_E) = \sum_{i \in rated,unrated} \left\{ \sum_{k=1}^{K} \frac{1}{2} \left[ -(\mu_{ki}^2 + \sigma_{ki}^2) + \log \sigma_{ki} \right] + \sum_{c=1}^{C} \sum_{\ell=1}^{\ell_c} a_{ic\ell} \log \hat{a}_{ic\ell} \right\}$$

**Derivation of Approximate Log-Likelihood.** We seek low-dimensional latent embeddings, $\vec{h}_i$. We marginalize $\vec{h}_i$ over the joint density in the second line of Equation A7. We expand this density to the predictive model and generative model as well as a prior over the product embedding.

$$\text{(A8)} \qquad \begin{aligned} &\log p(y_i, X_i | \vec{a}_i, \vec{\beta}) \\ &= \log \int p(y_i, X_i, \vec{h}_i | \vec{a}_i, \vec{\beta})\, \mathrm{d}\vec{h} \\ &= \log \int p_{pred}(y_i | \vec{h}_i, \vec{\beta}) p_{gen}(X_i | \vec{h}_i, \vec{\beta}) p_{prior}(\vec{h}_i | \vec{a}_i, \vec{\beta})\, \mathrm{d}\vec{h} \end{aligned}$$

We seek to learn an embedding *distribution* rather than just a point estimate of $\vec{h}_i$. We do not explicitly assume this form of the new joint density with the introduced product embedding, $\vec{h}_i$, and instead introduce a tractable distribution which we will use to approximate it, $q_{enc}(\vec{h}_i | X_i, \vec{a}_i, \vec{\beta})$, resulting in the "encoder model."

$$\text{(A9)} \qquad \begin{aligned} &\log \int p_{pred}(y_i | \vec{h}_i, \vec{\beta}) p_{gen}(X_i | \vec{h}_i, \vec{\beta}) p_{prior}(\vec{h}_i | \vec{a}_i, \vec{\beta})\, \mathrm{d}\vec{h} \\ &= \log \int \frac{p_{pred}(y_i | \vec{h}_i, \vec{\beta}) p_{gen}(X_i | \vec{h}_i, \vec{\beta}) p_{prior}(\vec{h}_i | \vec{a}_i, \vec{\beta})}{q_{enc}(\vec{h}_i | X_i, \vec{a}_i, \vec{\beta})} q_{enc}(\vec{h}_i | X_i, \vec{a}_i, \vec{\beta}) \mathrm{d}\vec{h} \\ &= \log \mathrm{E}_{\vec{h}_i \sim q_{enc}(\vec{h}_i | X_i, \vec{a}_i, \vec{\beta})} \left[ \frac{p_{pred}(y_i | \vec{h}_i, \vec{\beta}) p_{gen}(X_i | \vec{h}_i, \vec{\beta}) p_{prior}(\vec{h}_i | \vec{a}_i, \vec{\beta})}{q_{enc}(\vec{h}_i | X_i, \vec{a}_i, \vec{\beta})} \right] \end{aligned}$$

We find the best encoder model, $q_{enc}(\vec{h}_i | X_i, \vec{a}_i, \vec{\beta})$, for each datum $i$ from an assumed *family* of tractable densities. We estimate hyperparameters of the latent product embedding, $\vec{h}_i$, which index a unique element within an assumed variational distribution family.

Estimating these parameters using sampling techniques (e.g., MCMC) is intractable, hence we cast sampling as an optimization problem using a lower bound of the expectation via Jensen's inequality. This approximation is known as the "evidence lower bound," which is less than or equal to the intractable high-dimensional joint density, $\log p(y_i, X_i | \vec{a}_i, \vec{\beta})$.

$$\text{(A10)} \qquad \begin{aligned} &\log \mathrm{E}_{\vec{h}_i \sim q_{enc}(\vec{h}_i | X_i, \vec{a}_i, \vec{\beta})} \left[ \frac{p_{pred}(y_i | \vec{h}_i, \vec{\beta}) p_{gen}(X_i | \vec{h}_i, \vec{\beta}) p_{prior}(\vec{h}_i | \vec{a}_i, \vec{\beta})}{q_{enc}(\vec{h}_i | X_i, \vec{a}_i, \vec{\beta})} \right] \\ &\geq \mathrm{E}_{\vec{h}_i \sim q_{enc}(\vec{h}_i | X_i, \vec{a}_i, \vec{\beta})} \left[ \log \frac{p_{pred}(y_i | \vec{h}_i, \vec{\beta}) p_{gen}(X_i | \vec{h}_i, \vec{\beta}) p_{prior}(\vec{h}_i | \vec{a}_i, \vec{\beta})}{q_{enc}(\vec{h}_i | X_i, \vec{a}_i, \vec{\beta})} \right] \end{aligned}$$

With the logarithm moved inside the expectation, we decompose the joint density into three separate terms: the predictive model, the generative models, and the ratio of the encoder and prior model. Under the expectation of the encoder model, this last term is the Kullback-Leibler divergence between the encoder and the prior over the embedding, $D_{KL}(\log q_{enc} || p_{prior})$.

(A11)

$$
\begin{aligned}
&\mathrm{E}_{\vec{h}_i \sim q_{enc}(\vec{h}_i | X_i, \vec{a}_i, \vec{\beta})}\left[\log \frac{p_{pred}(y_i|\vec{h}_i,\vec{\beta})p_{gen}(X_i|\vec{h}_i,\vec{\beta})p_{prior}(\vec{h}_i|\vec{a}_i,\vec{\beta})}{q_{enc}(\vec{h}_i|X_i,\vec{a}_i,\vec{\beta})}\right] \\
&= \mathrm{E}_{\vec{h}_i \sim q_{enc}(\vec{h}_i | X_i, \vec{a}_i, \vec{\beta})}\left[\log p_{pred}(y_i|\vec{h}_i,\vec{\beta})\right] + \mathrm{E}_{\vec{h}_i \sim q_{enc}(\vec{h}_i | X_i, \vec{a}_i, \vec{\beta})}\left[\log p_{gen}(X_i|\vec{h}_i,\vec{\beta})\right] \\
&\qquad - \mathrm{E}_{\vec{h}_i \sim q_{enc}(\vec{h}_i | X_i, \vec{a}_i, \vec{\beta})}\left[\log \frac{q_{enc}(\vec{h}_i|X_i,\vec{a}_i,\vec{\beta})}{p_{prior}(\vec{h}_i|\vec{a}_i,\vec{\beta})}\right] \\
&= \mathrm{E}_{\vec{h}_i \sim q_{enc}(\vec{h}_i | X_i, \vec{a}_i, \vec{\beta})}\left[\log p_{pred}(y_i|\vec{h}_i,\vec{\beta})\right] + \mathrm{E}_{\vec{h}_i \sim q_{enc}(\vec{h}_i | X_i, \vec{a}_i, \vec{\beta})}\left[\log p_{gen}(X_i|\vec{h}_i,\vec{\beta})\right] \\
&\qquad - D_{KL}\left(\log q_{enc}(\vec{h}_i|X_i,\vec{a}_i,\vec{\beta}) || p_{prior}(\vec{h}_i|\vec{a}_i,\vec{\beta})\right) \\
&= \ell^i_{approx}(\vec{\beta})
\end{aligned}
$$

These three terms comprise the approximation, $\ell^i_{approx}(\vec{\beta})$, that we maximize. Since the Kullback-Leibler divergence term is negative, maximizing the overall approximation includes minimizing distributional dissimilarity between the posterior of the embedding given by the encoder model, $q_{enc}(\vec{h}_i|X_i,\vec{a}_i,\vec{\beta})$, and the distributional prior that we choose, $p_{prior}(\vec{h}_i|\vec{a}_i,\vec{\beta})$. If we minimize this divergence to zero, the approximate likelihood is equal to the true likelihood, i.e., $\log p(y_i, X_i|\vec{a}_i,\vec{\beta}) = \ell^i_{approx}(\vec{\beta})$. Thus, maximizing $\ell^i_{approx}(\vec{\beta})$ lower bounds the previously intractable likelihood maximization of $\log p(y_i, X_i|\vec{a}_i,\vec{\beta})$.

**Attributes.** Adding the latent variables for parameters of the multinomial attribute classifier, $\vec{\pi}_i$, results in a double integral and a corresponding expectation over the joint density of both $\vec{h}_i$ and $\vec{\pi}_i$. Our assumptions on factorization of the latent terms splits into two KL-divergence terms in the last line of the derivation. See Keng (2017) for additional discussion on the relation between KL-divergence and the cross-entropy loss term.

$$\log p\left(y_i, X_i|\vec{a}_i,\vec{\beta}\right)$$

$$= \log \iint p\left(y_i, X_i, \vec{h}_i, \vec{\pi}_i|\vec{a}_i,\vec{\beta}\right) \mathrm{d}\vec{h} d\vec{\pi}$$

$$= \log \iint p_{pred}\left(y_i|\vec{h}_i,\vec{\beta}\right) p_{gen}\left(X_i|\vec{h}_i,\vec{\beta}\right) p_{prior}\left(\vec{h}_i|\vec{a}_i,\vec{\beta}\right) p_{attr}\left(\vec{\pi}_i|\vec{a}_i,\vec{h}_i,\vec{\beta}\right) \mathrm{d}\vec{h} d\vec{\pi}$$

$$= \log \iint \frac{p_{pred}\left(y_i|\vec{h}_i,\vec{\beta}\right) p_{gen}\left(X_i|\vec{h}_i,\vec{\beta}\right) p_{prior}\left(\vec{h}_i|\vec{a}_i,\vec{\beta}\right) p_{attr}\left(\vec{\pi}_i|\vec{a}_i,\vec{h}_i,\vec{\beta}\right)}{q\left(\vec{h}_i,\vec{\pi}_i|\vec{a}_i, X_i,\vec{\beta}\right)} q\left(\vec{h}_i,\vec{\pi}_i|\vec{a}_i, X_i,\vec{\beta}\right) \mathrm{d}\vec{h} d\vec{\pi}$$

$$= \log \mathrm{E}_{\vec{h}_i,\vec{\pi}_i \sim q(\vec{h}_i,\vec{\pi}_i|\vec{a}_i,X_i,\vec{\beta})} \left[\frac{p_{pred}\left(y_i|\vec{h}_i,\vec{\beta}\right) p_{gen}\left(X_i|\vec{h}_i,\vec{\beta}\right) p_{prior}\left(\vec{h}_i|\vec{a}_i,\vec{\beta}\right) p_{attr}\left(\vec{\pi}_i|\vec{a}_i,\vec{h}_i,\vec{\beta}\right)}{q\left(\vec{h}_i,\vec{\pi}_i|\vec{a}_i, X_i,\vec{\beta}\right)}\right]$$

$$\geq \mathrm{E}_{\vec{h}_i,\vec{\pi}_i \sim q(\vec{h}_i,\vec{\pi}_i|\vec{a}_i,X_i,\vec{\beta})} \left[\log \frac{p_{pred}\left(y_i|\vec{h}_i,\vec{\beta}\right) p_{gen}\left(X_i|\vec{h}_i,\vec{\beta}\right) p_{prior}\left(\vec{h}_i|\vec{a}_i,\vec{\beta}\right) p_{attr}\left(\vec{\pi}_i|\vec{a}_i,\vec{h}_i,\vec{\beta}\right)}{q\left(\vec{h}_i,\vec{\pi}_i|\vec{a}_i, X_i,\vec{\beta}\right)}\right]$$

$$\begin{aligned} &= \mathrm{E}_{\vec{h}_i,\vec{\pi}_i \sim q(\vec{h}_i,\vec{\pi}_i|\vec{a}_i,X_i,\vec{\beta})} \Bigg[\log p_{pred}\left(y_i|\vec{h}_i,\vec{\beta}\right) + \log p_{gen}\left(X_i|\vec{h}_i,\vec{\beta}\right) + \log \frac{p_{prior}\left(\vec{h}_i|\vec{a}_i,\vec{\beta}\right)}{q_{enc}\left(\vec{h}_i|X_i,\vec{a}_i,\vec{\beta}\right)} \\ &\qquad + \log \frac{p_{attr}\left(\vec{\pi}_i|\vec{a}_i,\vec{\beta}\right)}{q_{attr}\left(\vec{\pi}_i|X_i,\vec{\beta}\right)}\Bigg] \end{aligned} \tag{A12}$$

$$\begin{aligned} &= \mathrm{E}_{\vec{\pi}_i \sim q(\vec{\pi}_i|X_i,\vec{\beta})} \Bigg[\mathrm{E}_{\vec{h}_i \sim q(\vec{h}_i|\vec{a}_i,X_i,\vec{\beta})} \Bigg[\log p_{pred}\left(y_i|\vec{h}_i,\vec{\beta}\right) + \log p_{gen}\left(X_i|\vec{h}_i,\vec{\beta}\right) \\ &\qquad - \log \frac{q_{enc}\left(\vec{h}_i|X_i,\vec{a}_i,\vec{\beta}\right)}{p_{prior}\left(\vec{h}_i|\vec{a}_i,\vec{\beta}\right)} - \log \frac{q_{attr}\left(\vec{\pi}_i|X_i,\vec{\beta}\right)}{p_{attr}\left(\vec{\pi}_i|\vec{a}_i,\vec{\beta}\right)}\Bigg]\Bigg] \end{aligned}$$

$$\begin{aligned} &= \mathrm{E}_{\vec{h}_i \sim q(\vec{h}_i|\vec{a}_i,X_i,\vec{\beta})} \left[\log p_{pred}\left(y_i|\vec{h}_i,\vec{\beta}\right)\right] + \mathrm{E}_{\vec{h}_i \sim q(\vec{h}_i|\vec{a}_i,X_i,\vec{\beta})} \left[\log p_{gen}\left(X_i|\vec{h}_i,\vec{\beta}\right)\right] \\ &\qquad - \mathrm{D}_{KL}\left[q(\vec{h}_i|\vec{a}_i, X_i,\vec{\beta}) || p_{prior}\left(\vec{h}_i|\vec{a}_i,\vec{\beta}\right)\right] - \mathrm{D}_{KL}\left(q_{attr}\left(\vec{\pi}_i|X_i,\vec{\beta}\right) | \, p_{attr}\left(\vec{\pi}_i|\vec{a}_i,\vec{\beta}\right)\right) \end{aligned}$$

### A.3. Gradient Backpropagation Using Local Reparameterization

We train the model by minimizing the loss functions in Table 1 with first-order stochastic gradient methods using mini-batches of training data. Specifically, we use the Adam stochastic gradient optimizer (Kingma and Ba 2015). Stochastic gradient methods are justified given their empirical performance and scalability via the backpropagation algorithm. Backpropagation simplifies an otherwise large calculation of a multi-parameter gradient to an equivalent series of smaller iterative gradient calculations. Gradients for a given loss in Table 1 propagate from the layer calculating the loss backwards to "earlier" layers, thereby taking advantage of the compositional structure of network layers and the chain rule of differentiation.

To use gradient methods, we employ the "reparameterization trick" used in Kingma and Welling (2013) and further popularized by the success of the variational autoencoder (VAE). We rewrite the otherwise intractable gradient of an expectation over the embedding to an equivalent tractable formulation by splitting the stochastic Gaussian embedding distribution into a deterministic neural net and an independent additive stochastic term. With this simplification it is feasible to compute an unbiased estimate of the gradient using Monte Carlo samples of the independent additive term. We similarly use this reparameterization trick when we do not have access to product attributes during training and inference. In this case, we use a relaxation of the otherwise non-differentiable categorical attribute variables called the Gumbel-Softmax (Jang, Gu, and Poole 2016).

### A.4. Example Masks for Compact Utility Vehicles (CUV)

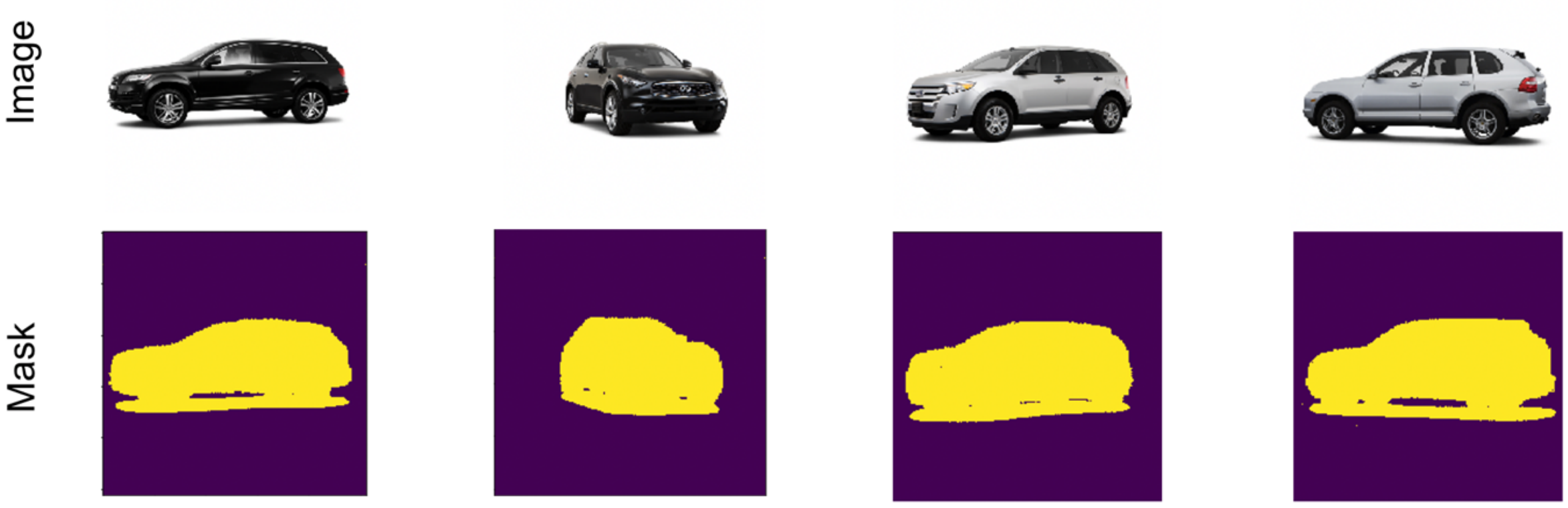

### A.5. Examples of Images at Successive Stages of Progressive Training.

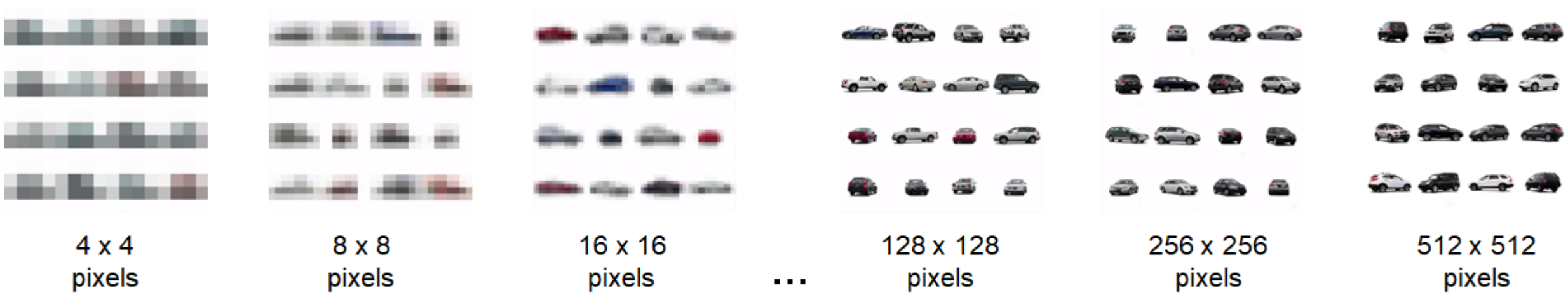

### A.6. Example Rating Page from Aesthetic Rating Survey used in Theme Clinic

SUV DESIGNER RATING PROGRESS: 1/10

RATE FROM 1 (VERY UNAPPEALING) TO 5 (VERY APPEALING)

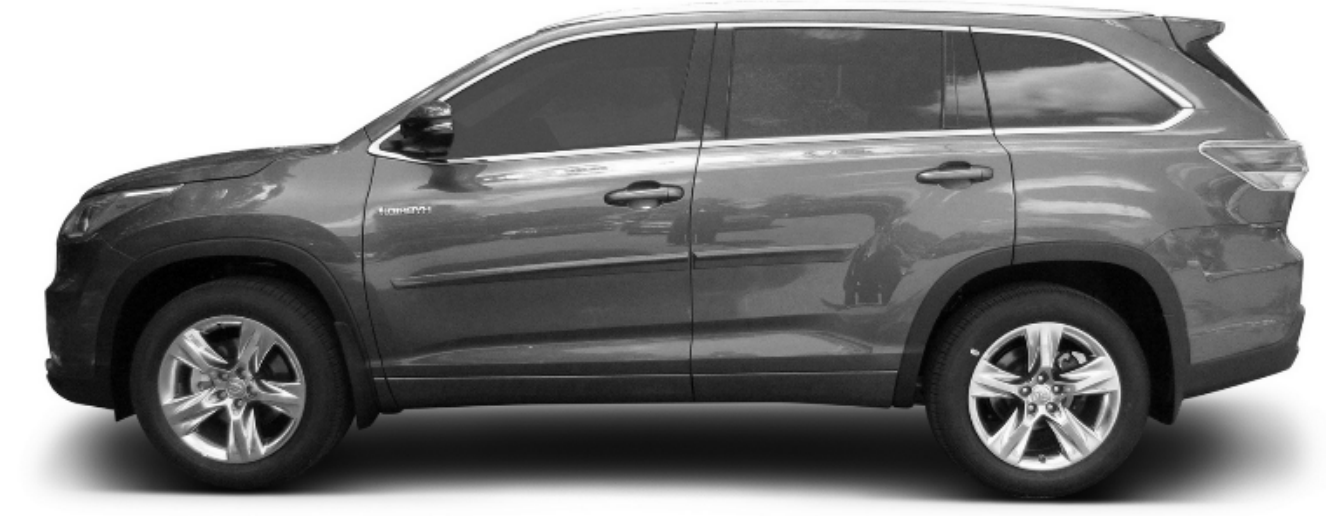

1 2 3 4 5

### A.7. Examples of Training Instabilities

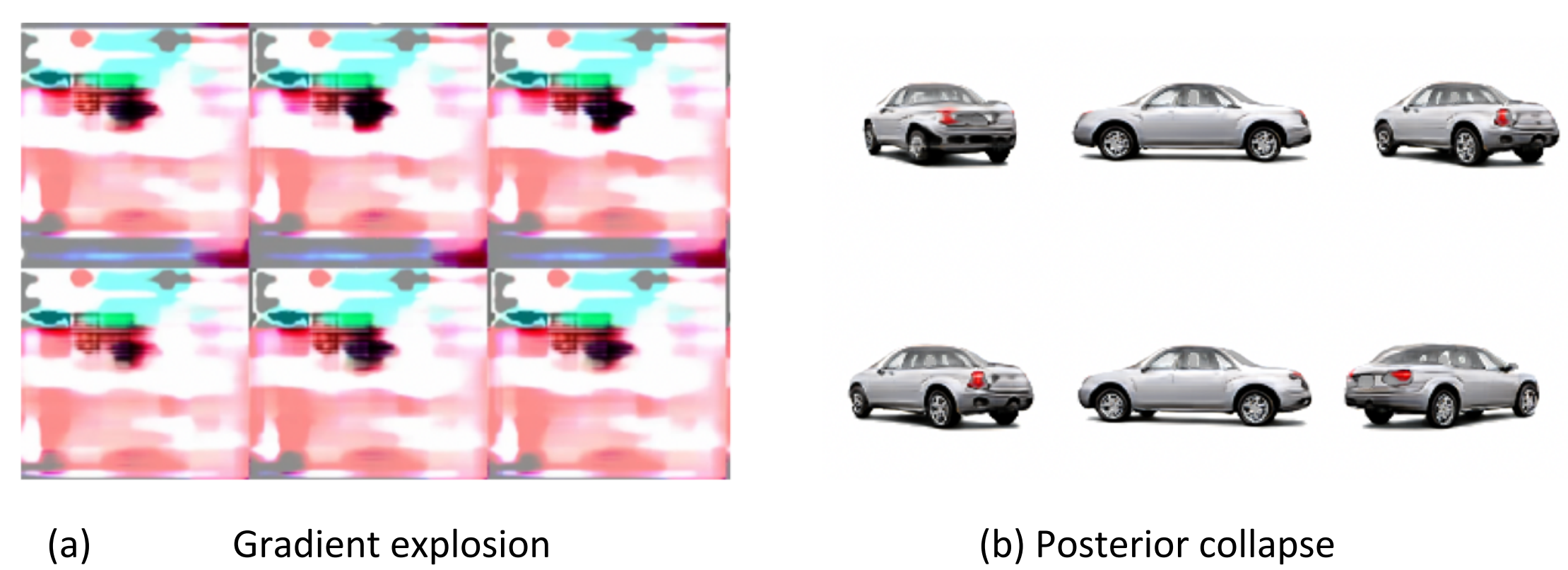

(a) Gradient explosion (b) Posterior collapse

### A.8. Embedding Dimensionality

We found that a 512-dimensional embedding was best able to encode information from images while still keeping the size of the entire model manageable to train. Underparameterization leads to insufficient model capacity, while overparameterization leads to excessively long training times. The following images provide examples of under- and overparameterization.

256-Dimensional Embedding

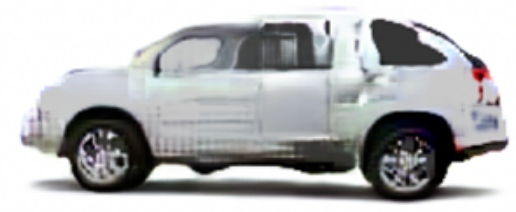
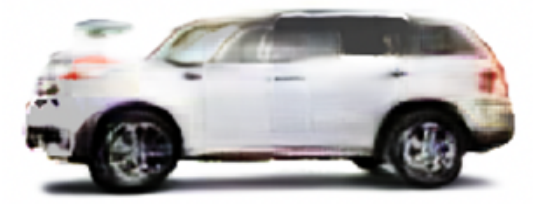
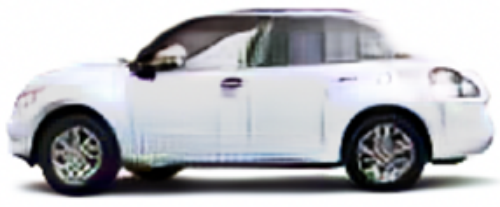

1024-Dimensional Embedding

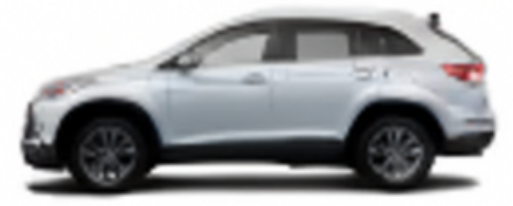
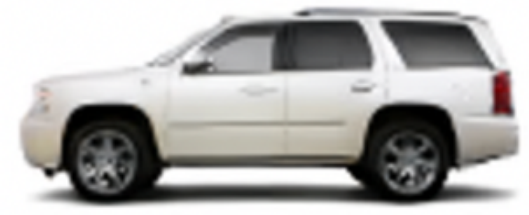
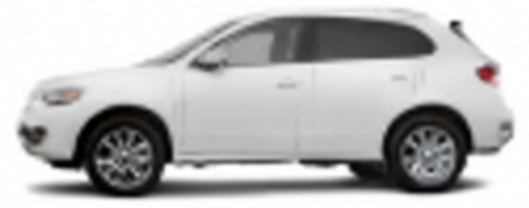

### A.9. Product Attributes Available for the Unlabeled Images

| Name | Description |
| --- | --- |
| Year | Model year of the vehicle (e.g., MY2014). Also known as “vintage.” |
| Brand | One of 48 possible brands in data. Each brand is often a subset of overall firm (e.g., Cadillac is a brand of General Motors). |
| Model | Indicator of vehicle in brand’s overall lineup (e.g., Audi A3 is smallest sedan offered by Audi in the U.S. market) |
| Viewpoint | Azimuth angle of the vehicle that image was taken from. |
| Body Type | Vehicle exterior categorization into one of: Convertible, Coupe, Sedan, Hatchback, Wagon, CUV, SUV, Truck, Minivan, Passenger Van, Cargo Van |
| Color | RGB coding of the primary exterior color of the vehicle. |

### A.10. Deep Pretrained Model With and Without Product Attributes

| Prediction Model | Mean Absolute Error (std. dev.) |
| --- | --- |
| Benchmark: Pretrained VGG16 and Fine-Tuned Final Layers. Without Product Attributes | 0.405<br>(0.039) |
| Benchmark: Pretrained VGG16 and Fine-Tuned Final Layers. With Product Attributes | 0.411<br>(0.050) |

**A.11. Effect of Less Labeled Data (Rated Images) for Semi-Supervised Prediction; Comparison to YOLO**

| Prediction Model | Mean Absolute Error (std. dev.) |
|---|---|
| Benchmark: Pretrained VGG16 and Fine-Tuned Final Layers. 100% Labeled Data | 0.405 (0.039) |
| Benchmark: Pretrained VGG16 and Fine-Tuned Final Layers. 50% Labeled Data | 0.452 (0.078) |
| Proposed Machine Learning Augmentation (Custom Deep Learning) 100% Labeled Data | 0.350 (0.043) |
| Proposed Machine Learning Augmentation (Custom Deep Learning) 50% Labeled Data | 0.404 (0.025) |
| Proposed Machine Learning Augmentation (Custom Deep Learning) 25% Labeled Data | 0.463 (0.045) |
| Proposed Machine Learning Augmentation (Custom Deep Learning) 10% Labeled Data | 0.551 (0.104) |
| You Only Look Once (YOLOv5) | 0.422 (0.016) |

**A.12. Effect of Less Unlabeled Data (Unrated Images) for Semi-Supervised Prediction**

| Prediction Model | Mean Absolute Error (std. dev.) |
|---|---|
| Proposed Machine Learning Augmentation (Custom Deep Learning) 100% Unlabeled Data | 0.350 (0.043) |
| Proposed Machine Learning Augmentation (Custom Deep Learning) 50% Unlabeled Data | 0.382 (0.041) |
| Proposed Machine Learning Augmentation (Custom Deep Learning) 10% Unlabeled Data | 0.381 (0.041) |

**A.13. Effect of Restricting Model Training to Single Viewpoint (Sideview)**

| Prediction Model | Mean Absolute Error (std. dev.) |
|---|---|
| Proposed Machine Learning Augmentation (Custom Deep Learning) All Viewpoints | 0.350 (0.043) |
| Proposed Machine Learning Augmentation (Custom Deep Learning) Single Viewpoint | 0.386 (0.036) |

#### A.14. Predictive Test for Aesthetic Innovativeness

We recalibrated the proposed model to predict aesthetic "innovativeness" using a previously-trained (proposed) model originally trained for aesthetic "appeal". We followed the same procedure as the pretrained VGG model (see §7.1.) by replacing the *predictor* for "appeal" with new (untrained) neural network layers to now predict "innovativeness". Likewise, we initially trained only the new layers while freezing the rest of the previously-trained model to allow the new layers' parameters to stabilize, followed by "finetuning" training of the entire model. The resulting predictive results for "innovativeness" are given below.

| Prediction Model | Mean Absolute Error (std. dev.) | Improvement |
|---|---|---|
| Baseline: Median Rating in Training Images (Constant Rating) | 0.627 (0.069) | 0.0 % |
| Benchmark: Computer Vision Features and Random Forest (Conventional Machine Learning) | 0.496 (0.064) | 20.9 % |
| Benchmark: VGG16 with Fine-Tuned Final Layers (Pretrained Deep Learning) | 0.311 (0.032) | 50.4 % |
| Proposed Machine Learning Augmentation (Custom Deep Learning) Fine-Tuned Final Layers | 0.253 (0.063) | 59.6 % |

#### A.15. Brief Definitions of Machine Learning Terms

2D average pooling. Partitions the input data (e.g., activations from previous neural net layer) over two spatial dimensions and computes the average value within each of the subsets.

2D convolution. Convolution is performed along two spatial dimensions of the input data. Convolution multiplies and accumulates from overlapping samples of the input data using learned kernels.

Adaptively balanced training losses. The loss function used for model training consists of several weighted terms which dynamically change in reaction to training stability metrics.

Annealed from zero. Increasing the value of a loss term's weighting coefficient as training progresses.

Batch normalization. Fix the means and variances of each layer's inputs to make the neural network faster and more stable.

Leaky rectified linear. The output function will be the input if the input is positive, otherwise the output will be a constant times the input. The constant is usually less than 1.

Lipschitz continuity. The absolute value of a gradient is constrained to be no larger than a constant along any given direction.

Minibatches. Splits of the training data into small subsets to calculate losses and update coefficients.

Neuronal receptive field. The subset of input from a previous layer that feed into a single "neuron" in a neural network, analogous to "neurons" in the human visual cortex V1 and V2.

Rectified linear. The output function will be the input if the input is positive, otherwise the output will be zero.

Residual connections. They allow gradients to flow through a network directly without passing through non-linear activation functions.

Spectral normalization. A weight normalization procedure that stabilizes the training of deep neural networks. Replace every weight in a layer's weight matrix with the weight divided by the largest eigenvalue of the matrix.

Squeeze-and-Excite. A neural network layer that explicitly models the interdependencies across channels (e.g., RGB for the input layer, number of kernels for convolutional layer) from the previous layer by 2D pooling the layer and performing "self-attention" to learn dependencies.

Stochastic gradient. A stochastic approximation to the gradient used in gradient descent optimization. Replaces the gradient with an estimate based on a subset of the data.

## Declarations

### Funding and Competing Interests

Alex Burnap received support from General Motors to partially fund a postdoctoral research position for the research conducted in this work. He certifies that none of the research or its results were censored or obfuscated in its publication. John Hauser and Artem Timoshenko certify that they have no affiliations with or involvement in any organization or entity with any financial interest or non-financial interest in the subject matter or materials discussed in this manuscript.